%% file: bare_adv.tex
\documentclass[10pt,journal,compsoc]{IEEEtran}
\ifCLASSOPTIONcompsoc
  \usepackage[nocompress]{cite}
  \usepackage[colorlinks=false,linkcolor=black,citecolor=black]{hyperref}
  \hypersetup{hidelinks} 
  \usepackage[utf8]{inputenc}
  \usepackage{graphicx}
  \usepackage{caption}
  \usepackage{multirow} 
  \usepackage{makecell} 
  \usepackage{pifont}
  \usepackage{textcomp}
  \usepackage[table,xcdraw]{xcolor}
  \usepackage{float}
  \usepackage{amsmath} 
  \usepackage{amsfonts}
  \usepackage{booktabs}
  \usepackage{subcaption}
  \usepackage{color}
  \usepackage{newtxmath}

  \usepackage{bbding}
  \usepackage{pifont}
  \usepackage{wasysym}
  \input{def}

\else
  \usepackage{cite}
\fi
\ifCLASSINFOpdf
\else
\fi
\definecolor{deepgreen}{RGB}{0, 100, 0} 

\begin{document}
%
\title{SKDF: A Simple Knowledge Distillation Framework for Distilling Open-Vocabulary Knowledge to Open-world Object Detector}
%
%
%
%

\author{Shuailei~Ma,~\IEEEmembership{}
        Yuefeng~Wang,~\IEEEmembership{}
        Ying~Wei,~\IEEEmembership{}
        Jiaqi~Fan,~\IEEEmembership{}
        Enming~Zhang,~\IEEEmembership{}
        Xinyu~Sun\IEEEmembership{}
        and~Peihao~Chen\IEEEmembership{}
\IEEEcompsocitemizethanks{\IEEEcompsocthanksitem Shuailei Ma, Yuefeng Wang, Jiaqi Fan, Enming Zhang are with College of Information Science and Engineering, Northeastern University, Shenyang, China, 110819.\protect\\
E-mail:\{xiaomabufei,wangyuefeng0203,f1074979751,z1693663290\} @gmail.com

\IEEEcompsocthanksitem Ying Wei is the corresponding author, with College of Information Science and Engineering, Northeastern University, Shenyang, China, 110819. \protect\\
E-mail: weiying@ise.neu.edu.cn

\IEEEcompsocthanksitem Xinyu Sun and Peihao Chen are with School of Software Engineering, South China University of Technology, China, 510641. \\
E-mail:\{csxinyusun,phchencs\} @gmail.com
}

}

%
%

\markboth{Journal of \LaTeX\ Class Files,~Vol.~14, No.~10, January~2024}%
{Shell \MakeLowercase{\textit{et al.}}: Bare Advanced Demo of IEEEtran.cls for IEEE Computer Society Journals}
%



\IEEEtitleabstractindextext{%
\begin{abstract}
Open World Object Detection (OWOD) is a novel computer vision task with a considerable challenge, bridging the gap between classic object detection (OD) benchmarks and real-world object detection. In addition to detecting and classifying seen/known objects, OWOD algorithms are expected to localize all potential unseen/unknown objects and incrementally learn them. The large pre-trained vision-language grounding models (VLM, \eg, GLIP) have rich knowledge about the open world, but are limited by text prompts and cannot localize indescribable objects. However, there are many detection scenarios in which pre-defined language descriptions are unavailable during inference. In this paper, we attempt to specialize the VLM model for OWOD tasks by distilling its open-world knowledge into a language-agnostic detector. Surprisingly, we observe that the combination of a simple \textbf{knowledge distillation} approach and the automatic pseudo-labeling mechanism in OWOD can achieve better performance for unknown object detection, even with a small amount of data. Unfortunately, knowledge distillation for unknown objects severely affects the learning of detectors with conventional structures for known objects, leading to catastrophic forgetting. To alleviate these problems, we propose the \textbf{down-weight loss function} for knowledge distillation from vision-language to single vision modality. Meanwhile, we propose the \textbf{cascade decouple decoding structure} that decouples the learning of localization and recognition to reduce the impact of category interactions of known and unknown objects on the localization learning process. Ablation experiments demonstrate that both of them are effective in mitigating the impact of open-world knowledge distillation on the learning of known objects. Additionally, to alleviate the current lack of comprehensive benchmarks for evaluating the ability of the open-world detector to detect unknown objects in the open world, we propose two benchmarks, which we name ``\textbf{StandardSet}$\heartsuit$'' and ``\textbf{IntensiveSet}$\spadesuit$'' respectively, based on the complexity of their testing scenarios. Comprehensive experiments performed on OWOD, MS-COCO, and our proposed benchmarks demonstrate the effectiveness of our methods. The code and proposed dataset are available at \url{https://github.com/xiaomabufei/SKDF}.

\end{abstract}

}

\maketitle

\IEEEdisplaynontitleabstractindextext

%
\IEEEpeerreviewmaketitle

\ifCLASSOPTIONcompsoc
\IEEEraisesectionheading{\section{Introduction}\label{sec:introduction}}
\else
\section{Introduction}
\label{sec:introduction}
\fi

%
%
%
%

\IEEEPARstart{O}{pen}-world object detection (OWOD) is a more practical detection problem in computer vision,  facilitating the development of object detection (OD) \cite{fasterrcnn,ddetr,detr,detection1,detection2,detection3,detection4,detection5,detection6,detection7,zou2019object} in the real world. Within the OWOD paradigm, the model’s lifespan is pushed by an iterative learning process, as shown in Fig. \ref{figure2} (a). At each episode, the model trained on the data with known object annotations is expected to detect known objects and all potential unknown objects. Human annotators then label a few of these tagged unknown classes of interest gradually. The model given these newly-added annotations continues incrementally updating its knowledge without retraining from scratch. 

\begin{figure}[htbp]
    \centering
    \includegraphics[width=0.9\linewidth]{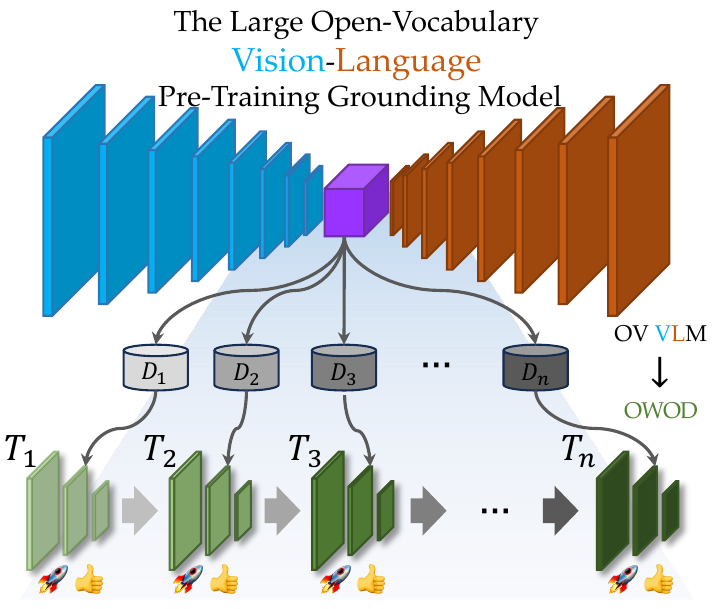}
    \caption{SKDF leverages the proposed down-weight training strategy to distill open-world knowledge from the large open-vocabulary pre-trainied vision-language model to the expert open-world detector with faster-detecting speed and better performance via small amounts of data.}
    \label{figure1}
\end{figure}

In the existing works \cite{ORE, owdetr, OCPL, two-branch, zohar2023prob, yu2023open, ma2023annealing, dong2023parallel,jamonnak2023ow, Revisiting, wang2023random, kim2022learning}, the open-world detectors are expected to know about the open world through several datasets \cite{mscoco,voc} with tiny scales. However, the annotations of these datasets are too few to provide adequate object attributes for the model. It is difficult for the model to achieve the ideal goal through these datasets. The large pre-trained vision-language grounding models (VL) \cite{glip,glipv2,mdetr,clip,detclip} have rich knowledge of the open world due to countless millions of parameters, huge open-world datasets, and training costs. 

However, their detection could not leave the participation of text prompts. Before detecting, the text prompt of all objects must be pre-listed to be detected, and the object whose text prompt was not listed could not be detected.  There are many detection scenarios in which pre-defined language descriptions are unavailable during inference. Therefore, it poses a challenge how to equip the VLM with language-agnostic unknown object detection capability. In addition, their detection speed is also a criticism due to the following question. $i)$ The huge number of parameters and FLOPs. $ii)$ The large pre-trained vision-language grounding model could only infer with several text prompts for the detecting performance, so they must infer many times when the number of prompts is large.

Humans' ability to recognize objects they have not seen before largely depends on their brains' knowledge base. Inspired by how humans face the open world, we propose to learn from the large pre-trained vision-language grounding models by knowledge distillation for OWOD. In this paper, we attempt to specialize the VLM model for OWOD tasks by distilling its open-world knowledge into a language-agnostic detector, as shown in Fig.\ref{figure1}. We were surprised to observe that the simple knowledge distillation approach for the OWOD algorithm can achieve better performance for unknown object detection than the large open-vocabulary pertaining vision-language model, even with a small amount of data. Unfortunately, knowledge distillation for unknown objects severely affects the learning of the detector with conventional structures for known objects due to the following challenges:

(\texttt{\uppercase\expandafter{\romannumeral1}}) Different from traditional Knowledge Distillation \cite{gou2021knowledge, park2019relational, tung2019similarity, mirzadeh2020improved,zhao2022decoupled}, the teacher and student contain different modalities and have different training manners, structures, and inference procedures. Meanwhile, the learning process of  OWOD algorithms \cite{ORE,owdetr,CAT,two-branch,OCPL,UC-OWOD} always has its own specific open-world pseudo-unknown labels. Therefore, the existing distilling objectives do not work and it is difficult to mix the learning of known grounding truth, unknown distilling knowledge and unknown pseudo-knowledge. The performance of detecting known objects is also crucial for OWOD. Through the experiments, we reveal that the direct use of distilled knowledge dramatically affects the model's learning ability of the original annotations. The model's performance in detecting known objects is damaged substantially. To alleviate this, we propose the down-weight training loss for the distillation training, which utilizes the distilled labels' object confidence from the large pre-trained vision-language grounding model and the searched pseudo objectness to reduce the weight of unknown loss in the total loss during training, as shown in Fig.\ref{figure2} (b).

(\texttt{\uppercase\expandafter{\romannumeral2}}) The presence of objects with highly similar features to known classes within the ``unknown objects'' can greatly affect the process of open-world object identification. In particular, the inclusion of unknown objects will impact the detector's performance on known objects. This issue impacts not only the identification process but also the localization process for models that use coupled information for both tasks. Therefore, we propose decoupling the detection learning process. However, the decoupled structure completely separates the localization and identification, leading to many potential problems (\eg mismatch between localization and identification results). Therefore, we propose a cascade structure that decouples the detecting process via two decoders and connects the two decoders via a cascade manner, as shown in Fig.\ref{figure2} (a). In this structure, foreground localization can be protected from category knowledge because the loss of identification is diluted by the latter decoder. Moreover, the identification process can utilize the localization information because it leverages the former decoder's output embeddings as the input queries.

\begin{figure*}[htbp]
  \centering
  \includegraphics[width = \textwidth]{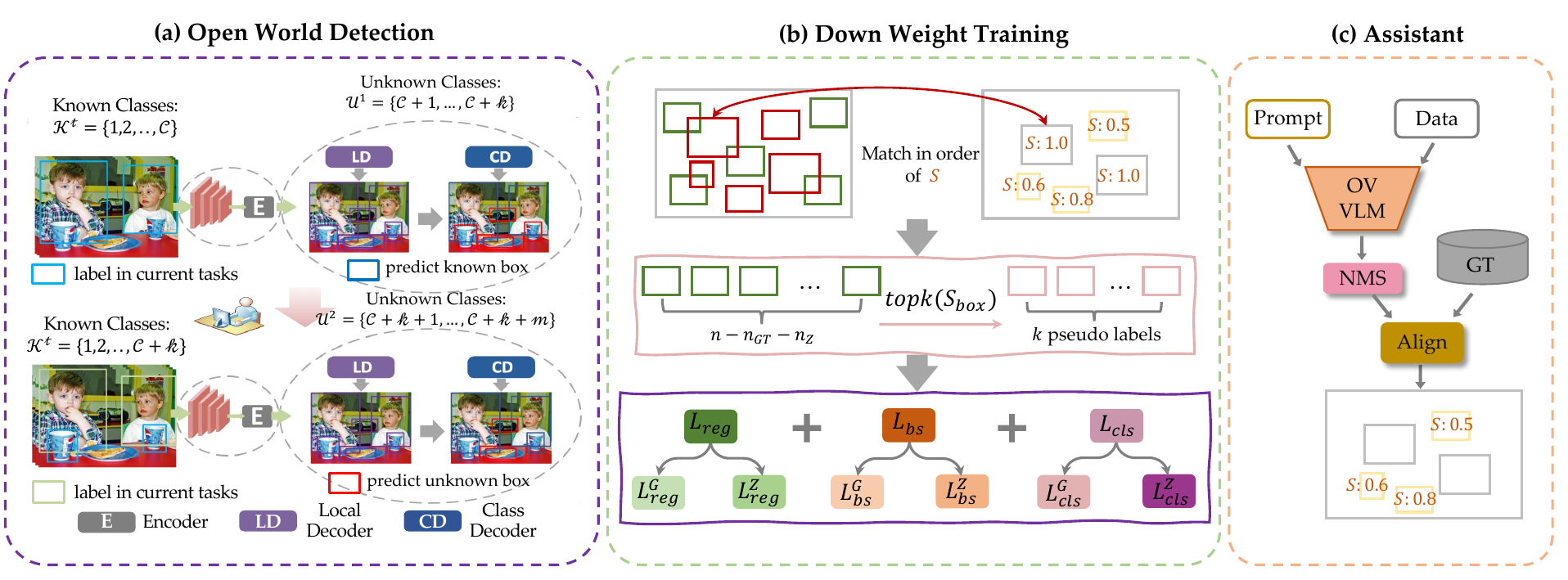}
  \caption{\textbf{Overall scheme of the proposed framework.} \texttt{(a)} illustrates the lifespan of the cascade open-world object detector where the model detects known objects and potential unknowns, with human annotators progressively labeling some unknown classes, the model incrementally updates its knowledge using these new labels without fully retraining. \texttt{(b)} exhibits the down-weight training strategy which leverages the objectness to separate the learning weight of the annotated known knowledge, distilled open-world knowledge, and searched pseudo-open-world unknown knowledge. \texttt{(c)} describes the distillation procedure that leverages the large-scale vocabulary prompt to mine the open-world knowledge in the open-vocabulary vision-language pertaining model.}
  \label{figure2}
\end{figure*}

Since existing datasets \cite{mscoco,voc} are manually annotated with predefined categories, current benchmarks cannot comprehensively measure the detection performance of open-world detectors for unknown objects due to the lack of bounding box annotations for unknown entities in the test scenarios. UnSniffer \cite{unsniffer} chooses testing images with only a few unknown foreground objects (no more than 3) from the MSCOCO dataset \cite{mscoco} to evaluate the detector's ability to identify unknown objects. Inspired by this, we propose two benchmarks, named \textbf{StandardSet}$\heartsuit$ and \textbf{IntensiveSet}$\spadesuit$ respectively, based on the complexity of their testing scenarios. Different from UnSiffer follows the manual annotation guidelines of the MSCOCO \cite{mscoco} dataset, inspired by the more detailed annotation criteria of finer-grained datasets \cite{lvis}, we manually select suitable evaluation scenes (with no fewer than three unknown objects) and provide more meticulous manual annotations of unknown objects. Moreover, our \textbf{IntensiveSet}$\spadesuit$, features an average of over 33 unknown open-world object instances per test scene. Our contributions can be summarized fourfold: 
\begin{itemize}\setlength{\itemsep}{2pt}
\item[$\bullet$] We observe that the simple \textbf{knowledge distillation} could convert the open-world knowledge in the large pre-trained vision-language grounding model for the specialized OWOD task and propose a simple framework with surprisingly good performance.
\item[$\bullet$]To mitigate the effect of distilled knowledge on the detection performance of known objects, we propose the \textbf{down-weight training loss function} for the detector's mixed learning process of known ground truth, distilled unknown knowledge, and the pseudo unknown knowledge in OWOD algorithm. Meanwhile, a \textbf{cascade decoupled detection transformer structure} is proposed to alleviate the influence caused by unknown objects on detecting known objects.
\item[$\bullet$]We propose two novel benchmarks to comprehensively evaluate the ability of the open-world detectors to detect unknown open-world objects, named \textbf{StandardSet}$\heartsuit$ and \textbf{IntensiveSet}$\spadesuit$ respectively, based on the complexity of their testing scenarios.
\item[$\bullet$] Our extensive experiments on existing and proposed benchmarks demonstrate the effectiveness of our framework. Our model exceeds the distilled large pre-trained vision-language grounding model for OWOD and state-of-the-art methods for OWOD and IOD. 
\end{itemize}

\section{Related Works}
\noindent\textbf{Large pre-trained vision-language models:}
Recently, inspired by the success of vision-language(VL) pre-trainied methods \cite{clip} and their good zero-shot ability, \cite{vild,glip,mdetr,glipv2,detclip} attempted to perform zero-shot detection on a larger range of domains by using pre-trained vision language models. \cite{vild} proposed a zero-shot detection method to distill knowledge from a pre-trained vision language image classification model. \cite{glip} tried to align region and language features using a dot-product operation and could be trained end-to-end on grounding and detection data. \cite{mdetr} proposed an end-to-end modulated detector that detects objects in an image conditioned on a raw text query, like a caption or a question. \cite{detclip} proposed a paralleled visual-concept pre-trainied method for open-world detection by resorting to knowledge enrichment from a designed concept dictionary. However, VLM requires predefined object categories to drive the model, therefore it cannot be directly applied to the OWOD tasks. Moreover, VLM requires substantial computational resources and a long runtime.

\par
\noindent\textbf{Semi-Supervised Learning For Object Detection:} In this area, there are two dominant approaches, the consistency methods \cite{ssl1,ssl2} and pseudo-label methods \cite{simplessl,etessl,ssl4,ssl3}, respectively. STAC \cite{simplessl} deploys highly confident pseudo labels of localized objects from an unlabeled image and updates the model by enforcing consistency via strong augmentations. Xu $et$ $al$.\cite{etessl} proposed an end-to-end pseudo-labeling framework to avoid the complicated training process and also achieve better performance. Liu $et$ $al$.\cite{ssl4} improved the pseudo-label generation model via teacher-student mutual learning regimen and addressed the crucial imbalance issue in generated pseudo-labels. However, these distillation methods are applicable to single-modality closed-set object detection tasks. This paper introduces a simple framework for distilling from a multimodal open-vocabulary model to a single-modality open-world model.
\par

\begin{figure*}[htbp]
  \centering
  \includegraphics[width = \textwidth]{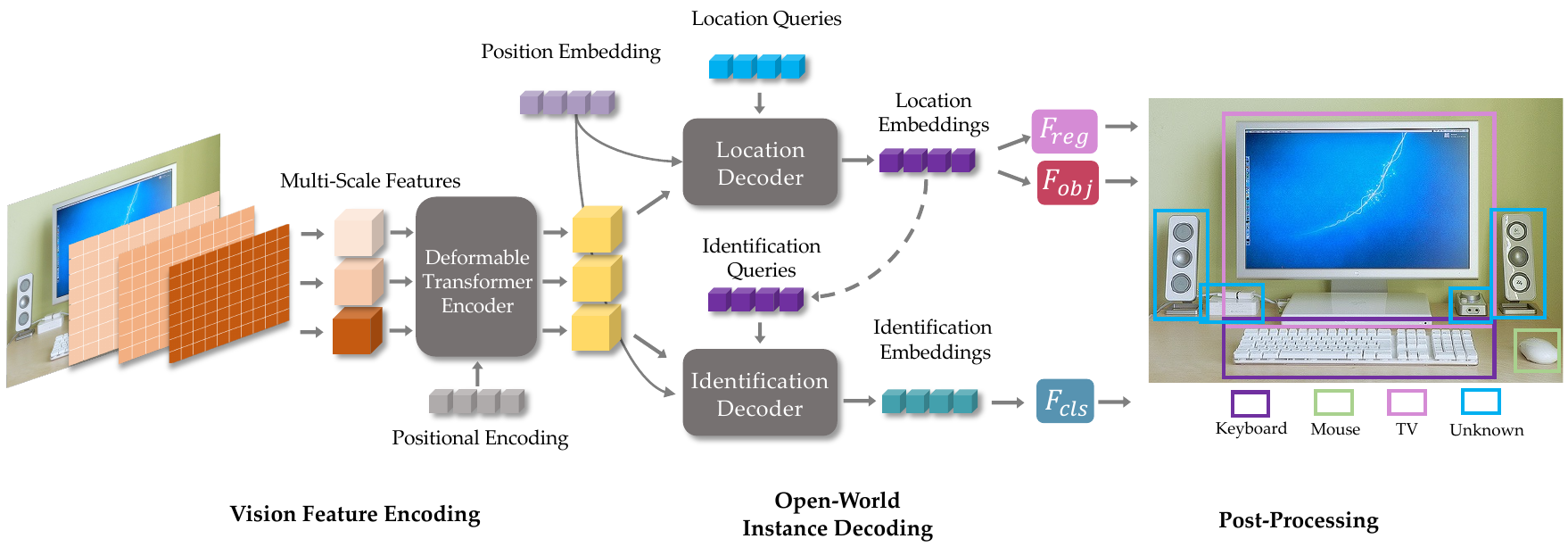}
  \caption{\textbf{Overall Architecture of proposed cascade decoupled open-world detector.} The proposed detector consists of a multi-scale feature extractor, the decoupled cascade transformer decoder, and the regression prediction branch. The multi-scale feature extractor comprises the mainstream feature extraction backbone and a deformable transformer encoder, for extracting multi-scale features. The decoupled cascade transformer decoders are the deformable transformer decoders and decouple the localization and identification process in the cascade way. The regression prediction branch contains the bounding box regression branch $F_{reg}$, novelty objectness branch $F_{obj}$, and novelty classification branch $F_{cls}$. The novelty classification and objectness branches are single-layer feed-forward networks (FFN) and the regression branch is a 3-layer FFN.}
  \label{figure3}
\end{figure*}

\noindent\textbf{Open-World Object Detection (OWOD):}
ORE \cite{ORE} introduced OWOD task and ORE which adapted the faster-RCNN model with feature-space contrastive clustering, an RPN-based unknown detector, and an Energy Based Unknown Identifier (EBUI) for the OWOD objective. Recently, several works \cite{OCPL,two-branch,UC-OWOD} attempted to extend ORE. OCPL \cite{OCPL} was proposed to learn the discriminative embeddings of known classes in the feature space to minimize the overlapping distributions of known and unknown classes. 2B-OCD \cite{two-branch} proposed a two-branch objectness-centric open-world object detection framework consisting of the bias-guided detector and the objectness-centric calibrator. OWDETR \cite{owdetr} proposed to utilize the pseudo-labeling scheme to supervise unknown object detection, where unmatched object proposals with high backbone activation are selected as unknown objects. Existing methods limit the training of the model to a small subset of object annotations, which fails to fully teach the model how to recognize objects and foreground. In this paper, we propose distilling open-world knowledge from large pre-trained models to scale up the open-world knowledge in the training data. Experiments prove that even the simple distillation framework can stimulate the model's ability to recognize unknown objects from the background even without the need for vast amounts of data, even achieving capabilities that surpass those of the teacher.

\section{Problem Formulation}
$\mathcal{K}^{t}=\{1,2, \ldots, C\}$ denote the set of known object classes and $\mathcal{U}^{t}=\{C+1, \ldots\}$ denote the unknown classes which might be encountered at the test time, at the time $t$. We labeled the known object categories $\mathcal{K}^{t}$ in the dataset $\mathcal{D}^{t}=\{\mathcal{J}^{t}, \mathcal{L}^{t}\}$ where $\mathcal{J}^{t}$ denotes the input images and $\mathcal{L}^{t}$ denotes the corresponding labels at time $t$. The training image set consists of $M$ images $\mathcal{J}^{t}=\{i_{1}, i_{2}, \ldots, i_{M}\}$ and corresponding labels $\mathcal{L}^{t}=\{\ell_{1}, \ell_{2}, \ldots, \ell_{M}\}$. Each $\ell_{i}=\{\mathcal{T}_{1}, \mathcal{T}_{2}, \ldots, \mathcal{T}_{N}\}$ denotes a set of $N$ object instances with their class labels $c_{n} \subset \mathcal{K}^{t}$ and locations, $\{ x_{n}, y_{n}, w_{n}, h_{n}\}$ denote the bounding box center coordinates, width and height respectively. 

The artificial assumptions and restrictions in closed-set object detection are removed in Open-World Object Detection. It aligns object detection tasks more with real life. It requires the trained model $\mathcal{M}_{t}$ to detect the previously encountered known classes $C$ and identify an unseen class instance as belonging to the unknown class. In addition, it requires the object detector to be capable of incremental updates for new knowledge, and this cycle continues over the detector’s lifespan. In the incremental updating phase, the unknown instances identified by $\mathcal{M}_{t}$ are annotated manually. Along with their corresponding training examples, they update $\mathcal{D}^{t}$ to $\mathcal{D}^{t+1}$ and $\mathcal{K}^{t}$ to $\mathcal{K}^{t+1}=\{1,2,\ldots, C,\ldots, C+\text{n}\}$. The model adds the $n$ new classes to known classes and updates itself to $\mathcal{M}_{t+1}$ without retraining from scratch on the whole dataset $\mathcal{D}^{t+1}$.

\section{Proposed method}
This section elaborates on the proposed framework in detail. We start by illustrating the overall scheme of our proposed framework in Sec.\ref{4.1} Furthermore, we introduce the open-world object detector in Sec.\ref{4.2}, the distillation process in Sec.\ref{4.3}, and the matching and pseudo-labeling procedure of the owod algorithm in Sec.\ref{4.4}. Last but not least, we describe the down-weight training strategy in Sec.\ref{4.5} and the inference phrase in Sec.\ref{4.6}.

\subsection{Overall Scheme} \label{4.1}
Fig.\ref{figure2} illustrates the overall scheme of our framework. For a given image $x \in \mathbb{R}^{H \times W \times3}$, it is first sent into the open-world detector and the large pre-trained vision-language grounding model simultaneously. The detector leverages the visual features of the input to predict the localization, box score, and classification. The large pre-trained vision-language grounding model mines unknown open-world knowledge from the inputs. The known ground truth and unknown distilled knowledge aggregate the open-world supervision. In the training phase, we match the prediction and open-world supervision according to the regression loss, classification, and supervision confidence. After matching, the pseudo labels are selected according to the predicted box score. Then all labels are leveraged to train the open-world detector by the down-weight training loss function. In addition, when new categories are introduced at each episode, based on an exemplar replay-based finetuning to alleviate the catastrophic forgetting of learned classes and the finetuning by using a balanced set of exemplars stored for all known classes, our detector could continuously learn during its lifespan.

\begin{figure*}[htbp]
  \centering
  \includegraphics[width = \textwidth]{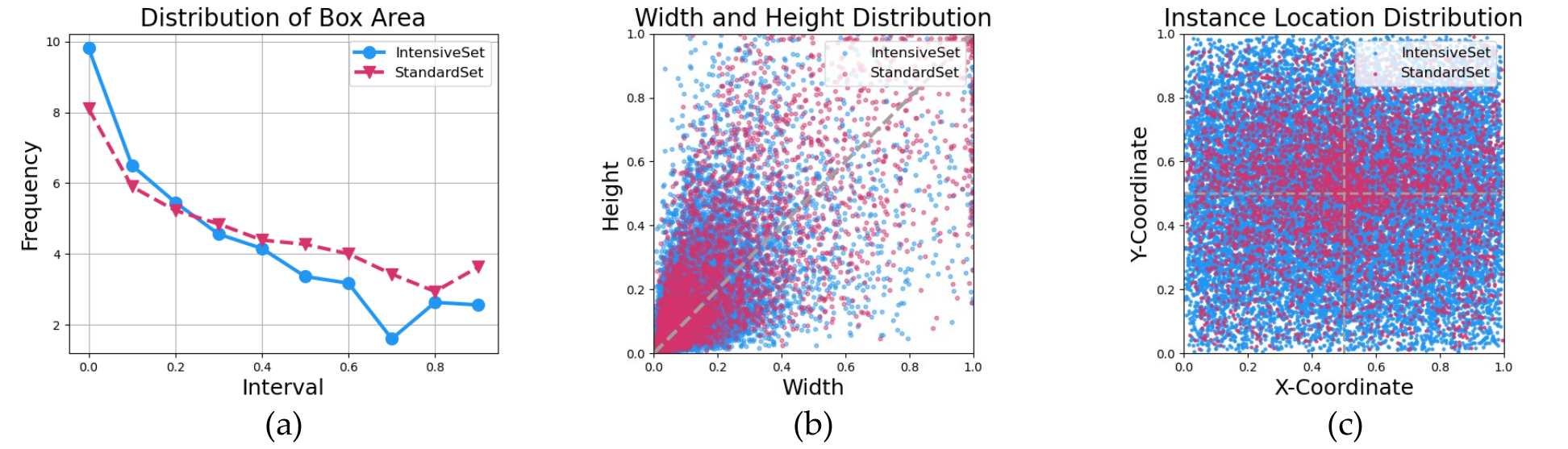}
  \caption{\textbf{The detailed data analysis of} \textbf{StandardSet}$\heartsuit$ \textbf{and} \textbf{IntensiveSet}$\spadesuit$. In (a), we calculate the area distribution of instances in the two benchmark test scenes, with the vertical axis representing the logarithm of the count with respect to Euler's number \texttt{e}. In (b) and (c), we respectively analyze the aspect ratio and the spatial distribution of the instance bounding box annotations.
  }
  \label{figure4}
\end{figure*}

\subsection{Open-World Object Detector}\label{4.2}
As shown in Fig.\ref{figure3}, the open-world detector first uses a hierarchical feature extraction backbone to extract multi-scale features $\mathrm{Z}_{i}$. The feature maps ${Z}_{i}$ are projected from dimension $C_s$ to dimension $C_d$ by using 1×1 convolution and concatenated to $N_s$ vectors with $C_d$ dimensions after flattening out. Afterward, along with supplement positional encoding $P \in \mathbb{R}^{N_{s} \times C_{d}}$, the multi-scale features are sent into the deformable transformer encoder to encode semantic features. The encoded semantic features are acquired and sent into the localization decoder together with a set of $N$ learnable location queries. 

Aided by interleaved cross-attention and self-attention modules, the localization decoder transforms the location queries $Q_{Location} \in \mathbb{R}^{N \times D}$ to a set of N location query embeddings $E_{Location} \in \mathbb{R}^{N \times D}$.  Meanwhile, the $E_{Location} $ are used as class queries and sent into the identification decoder together with the $M$ (\ie the encoded multi-scale features) again. The identification decoder transforms the class queries to $N$ class query embeddings $E_{class}$ that correspond to the location query embeddings. The operation of cascade decoders is expressed as follows:
\begin{equation}
\resizebox{.70\hsize}{!}{$E_{Location}=F_{LD}(F_E(\varnothing(x), P), Q_{Location}, R)$},
\end{equation}
\begin{equation}
\resizebox{.65\hsize}{!}{$E_{Class}=F_{ID}(F_E(\varnothing(x), P), E_{Location}, R)$}.
\end{equation}
where $F_{LD}(\cdot)$ and $F_{ID}(\cdot)$ denote the localization and identification decoder. $F_{E}(\cdot)$ is the encoder and $\varnothing(\cdot)$ is the backbone. $R$ represents the reference points and $x$ denotes the input image. $Q_{Class}$ stands for the class queries. Eventually, the $E_{Class}$ are then sent into the classification branch to predict the category $cls \in [0,1]^{N_{obj}}$. The $E_{Location} $ are then input to the regression and box score branch to locate N foreground bounding boxes $b \in [0,1]^{4}$ and predict the box score $bs \in [0,1]$. 

In this decoupling structure, foreground localization can be protected from category knowledge, and the identification process can utilize the localization information. Therefore, we alleviate the infusion of the unknown objects on the detecting performance of known objects and the confusion between the category and location of the same objects.

\subsection{Open-World Object Supervision} \label{4.3}
In the knowledge distillation phase, the teacher leverages a large pre-trained vision-language grounding model and a text prompt that contains object categories as many as possible to mine unknown open-world knowledge from input $x$. In this paper, we utilize GLIP \cite{glip} and categories of LVIS \cite{lvis} as the large pre-trained vision-language grounding model and the text prompts, respectively. The distilled open-world knowledge is first processed through the NMS produce. Then we align the distilled open-world knowledge and ground truth by the align module, where we align the generated labels to the given data annotation space \eg translate LVIS categories (trailer truck, tow truck, \etal) into COCO category truck and exclude the known set in the distilled open-world knowledge. In addition, the identification confidence of the unknown labels from the large pre-trained vision-language grounding model is reserved for the following training process and leveraged as the supervision confidence.

\subsection{Matching and Evolving} \label{4.4}
Following the existing open-world detectors \cite{ORE,owdetr,CAT,two-branch,OCPL,UC-OWOD}, we set a box score prediction branch that leverages the predicted box score to automatically select pseudo-unknown labels from the remaining regression boxes after the matching process of each training iteration as the automatic pseudo-labeling mechanism. When matching the open-world supervision and prediction, we consider the box regression loss, prediction class score, and confidence of supervision. 

After matching, we leverage the box score branch to help the detector learn more unknown open-world objects beyond the ground truth and distilled knowledge from the large pre-trained vision-language grounding model during training via selecting pseudo labels. We denote by $\hat{y}$ the open-world supervision, and $y={\{y_i\}}_{i=1}^N$ the set of $N$ predictions. For finding the best bipartite matching between them, a permutation of $N$ elements $\sigma \in \mathfrak{S}_{N}$ with the lowest cost is searched for as follows:
\begin{equation}
\hat{\omega}=\underset{\omega \in \mathfrak{S}_N}{\arg \min } \sum_i^N \mathcal{L}_{\operatorname{match}}(\boldsymbol{\hat{y}_i}, \boldsymbol{y_{\omega(i)}}),
\end{equation}
where $\mathcal{L}_{\text {match }}\left(\hat{y}_i, y_{\omega(i)}\right)$ is the pair-wise matching cost between $\hat{y}_i$ which represents ground truth or labels from the large pre-trained vision-language grounding model and a prediction $y$ with index $\omega(i)$, shown as Equation.\ref{match}. Inspired by \cite{ddetr,detr,mdetr}, we choose the Hungarian algorithm as the optimal assignment.
\begin{equation}\label{match}
\mathcal{L}_{\operatorname{match}}(\hat{y}_i, {y}_{\omega(i)}) = L_r(\boldsymbol{\hat{b}_i}, \boldsymbol{{b}_{\omega(i)}})- \boldsymbol{cls}_{\omega(i)}(\hat{c}_i)-\boldsymbol{\hat{S}_i},
\end{equation}
where $L_r$ denotes the regression loss, which consists of box loss and GIOU loss \cite{GIOU}. $\hat{b}$ and $b$ represent the open-world supervision box and prediction box, respectively. $\hat{S}$ is the confidence of the open-world supervision. Then, the pseudo labels are selected as follows:
\begin{equation}
l_p =\operatorname{Topk}(\{\boldsymbol{bs}_i\}_{i \notin \widehat{\omega}}),
\end{equation}
where $bs$ denotes the prediction box score. The pseudo-labels could prevent the model from falling entirely into the knowledge of the large pre-trained vision-language grounding model and help it know unseen objects beyond the distilled knowledge.

\subsection{Down-Weight Training Strategy} \label{4.5}
The inclusion of distilled knowledge inevitably influences the model's learning of the known set. Because the quality of it could not be guaranteed, and it increases the difficulty of detector learning. Therefore, we propose the down-weight training strategy, which leverages the distilled knowledge identification confidence to generate harmonic factor and down-weight the unknown training loss and train the detector in an end-to-end manner as shown in Fig.\ref{figure2} (b). The training loss function is as follows:
\begin{equation}
L=L_{r}+L_{bs}+L_{cls}+L_{r}^{kd}+L_{bs}^{kd}+L_{cls}^{kd}+L_{cls}^p, 
\end{equation}
where the $L_{r}$ uses the common regression loss which consists of box and GIOU loss \cite{GIOU}. $L_{bs}$ and $L_{cls}$ represent the box score and classification loss, respectively. They all leverage the common sigmoid focal loss \cite{focal}. For simplicity, we omit them. In addition, the $L_{r}^{kd}$, $L_{bs}^{kd}$, $L_{cls}^{kd}$ and $L_{cls}^p$ all utilize correspondingly down-weight loss function we propose, the formulations are shown as follows:
\begin{equation}
L_{r}^{kd}=\frac{1}{|\boldsymbol{l_{kd}}|} \sum_{i=1}^{N_q} \mathbf{}{1}_{\{i \in \boldsymbol{l_{kd}}\}}\hat{\boldsymbol{S}}_{\hat{\omega}(i)}[\|\boldsymbol{b}_i-\hat{\boldsymbol{b}}_{\hat{\omega}(i)}\|_1+1-\mathcal{G}(\boldsymbol{b}_i, \hat{\boldsymbol{b}}_{\hat{\omega}(i)})],
\end{equation}
\begin{equation}
L_{bs}^{kd}=\frac{1}{\sum_{i=1}^{N_q} \mathbf{}{1}_{\{i \in \boldsymbol{l_{kd}}\}}\|\boldsymbol{bs}_i\|_1} \sum_{i=1}^{N_q}\mathbf{1}_{\{i \in \boldsymbol{l_{kd}}\}}[l_{sf}(\boldsymbol{bs}_i, \hat{\boldsymbol{S}}_{\hat{\boldsymbol{\omega}}(i)})],
\end{equation}
\begin{equation}
L_{cls}^{kd}=\frac{1}{\sum_{i=1}^{N_q} \mathbf{1}_{\{i \in \boldsymbol{l_{kd}}\}}\|\boldsymbol{cls}_i\|_1} \sum_{i=1}^{N_q}\mathbf{1}_{\{i \in \boldsymbol{l_z}\}}[l_{sf}(\boldsymbol{cls}_i, \hat{\boldsymbol{S}}_{\hat{\boldsymbol{\omega}}(i)})],
\end{equation}
\begin{equation}
L_{cls}^p=\frac{1}{\sum_{i=1}^{N_q} \mathbf{1}_{\{i \in \boldsymbol{l_p}\}}\|\boldsymbol{cls}_i\|_1} \sum_{i=1}^{N_q}\mathbf{1}_{\{i \in \boldsymbol{l_p}\}}[l_{sf}(\boldsymbol{cls}_i, \boldsymbol{bs}_i)],
\end{equation}
where $\boldsymbol{l_{kd}}$ and $\boldsymbol{l_{p}}$ denote the distilled knowledge supervision and pseudo supervision labels, respectively. $N_q$ represents the number of queries, $\hat{\omega}(i)$ represents the index of the label corresponding to the prediction. $l_{sf}$ denotes the sigmoid focal loss function. $\mathcal{G}(\cdot)$ represents the GIOU loss function. $\hat{S}$ is the confidence of supervision labels. $b$, $bs$, and $cls$ are the prediction box, box score, and classification score, respectively.

\subsection{Inference}\label{4.6}
 During inference, our detector only utilizes the visual features of the inputs to detect open-world objects without any information from the other modalities. The inference process is to composite the detector output to form open-world object instances. Formally, the $i$-th output prediction is generated as $<b_i,\ bs_i,\ cls_i>$. According to the formulation : $s_i, l_i = max(cls_i)$, the result is acquired as $<l_i,\ s_i,\ b_i>$, where $l$ is the category, $s$ denotes the confidence, and $b$ represents the predicted bounding boxes.

\begin{table}[H]
\centering
\resizebox{0.95\linewidth}{!}{
\begin{tabular}{c|ccc}
\toprule \toprule
Dataset         & Images & Avg. Kn. & Avg.Un \\ \midrule
\texttt{StandardSet}$\heartsuit$     & 694    & 2.7      & 3.4    \\
\texttt{IntensiveSet} $\spadesuit$    & 489    & 5.8      & 33.7   \\ \bottomrule \bottomrule
\end{tabular}}
\caption{\textbf{The statistics on the number of scenes and instances} for the proposed testing sets: \textbf{StandardSet}$\heartsuit$ and \textbf{IntensiveSet}$\spadesuit$.}
\label{table5}
\end{table}

\section{Proposed Dataset}
Since existing datasets \cite{mscoco,voc} are manually annotated with predefined categories, current benchmarks cannot comprehensively measure the detection performance of open-world detectors for unknown objects due to the lack of bounding box annotations for unknown entities in the test scenarios.  In this paper, we propose two benchmarks, named \textbf{StandardSet}$\heartsuit$ and \textbf{IntensiveSet}$\spadesuit$ respectively, based on the complexity of their testing scenarios. For the proposed benchmark, the known categories are defined as the PASCAL VOC \cite{voc} dataset, and we train the open-world detectors by the images and annotations of the PASCAL VOC \cite{voc} training set. Meanwhile, inspired by the detailed annotation criteria of finer-grained datasets \cite{lvis}, we manually select suitable evaluation scenes (with no fewer than five unknown objects) and provide more meticulous manual annotations of unknown objects to construct the testing set, as shown in TABLE.\ref{table5}.

\subsection{Standard Testing Set}
The \textbf{StandardSet}$\heartsuit$ totally contains 694 testing scenes which we select from the MS-COCO \cite{mscoco} dataset. We manually annotate the unknown objects contained within. It contains 1897 known instances and 2378 unknown open-world instances with respect to the predefined known categories. We statistic the instance area, aspect radio, and position distribution in Fig.\ref{figure4}. The statistical results suggest that compared to \textbf{IntensiveSet}$\spadesuit$, \textbf{StandardSet}$\heartsuit$ has a more uniform distribution of instance areas. Additionally, the aspect ratio of bounding boxes in the \textbf{StandardSet}$\heartsuit$ test set tends towards a uniform distribution, with the majority of bounding box annotations concentrated in the central region of the scenes.

\subsection{Intensive Testing Set}
In the \textbf{IntensiveSet}$\spadesuit$ testing set, we elevate the complexity of the scenes by selecting 489 highly complex scenes, which altogether encompass 2859 annotations of known objects and 16482 bounding box annotations of unknown open-world objects that we manually annotate, averaging 5.8 annotations of known objects and 33.7 annotations of unknown objects per image. As shown in Fig.\ref{figure4}, compared to the \textbf{StandardSet}$\heartsuit$ testing set, the bounding box annotations in the \textbf{IntensiveSet}$\spadesuit$ testing set primarily consist of smaller areas, with a more uniform distribution in terms of aspect ratios and spatial positioning.

\begin{table}[htbp]
\centering
\renewcommand\arraystretch{1.25}
\resizebox{\linewidth}{!}{
\begin{tabular}{ccccc}
\toprule
\multicolumn{5}{c}{ OWOD split}\\\midrule \midrule
 Task ID $\rightarrow$& \textbf{Task 1}   & \textbf{Task 2 } & \textbf{Task 3}     & \textbf{Task 4}   \\ \midrule 
\multicolumn{1}{l|}{ Semantic split $\rightarrow$}      & \multicolumn{1}{c}{\begin{tabular}[c]{@{}c@{}}VOC \\ Classes\end{tabular}} & \multicolumn{1}{c}{\begin{tabular}[c]{@{}c@{}}Outdoor, \\ Accessories, \\ Appliances, \\ Truck\end{tabular}} & \multicolumn{1}{c}{\begin{tabular}[c]{@{}c@{}}Sports, \\ Food\end{tabular}} & \begin{tabular}[c]{@{}c@{}}Electronic, \\ Indoor, \\ Kitchen, \\ Furniture\end{tabular} \\ \midrule
\multicolumn{1}{l|}{$\#$ training images} & \multicolumn{1}{c}{16551}    & \multicolumn{1}{c}{45520}   & \multicolumn{1}{c}{39402}  & 40260    \\
\multicolumn{1}{l|}{$\#$ test images}     & \multicolumn{1}{c}{4952}    & \multicolumn{1}{c}{1914}   & \multicolumn{1}{c}{1642}    & 1738   \\
\multicolumn{1}{l|}{$\#$ train instances} & \multicolumn{1}{c}{47223}   & \multicolumn{1}{c}{113741}   & \multicolumn{1}{c}{114452}   & 138996      \\
\multicolumn{1}{l|}{$\#$ test instances}  & \multicolumn{1}{c}{14976}   & \multicolumn{1}{c}{4966}   & \multicolumn{1}{c}{4826}  & 6039  \\  \midrule\midrule
\multicolumn{5}{c}{ MS-COCO split} \\\midrule \midrule
 Task ID $\rightarrow$& \textbf{Task 1}   & \textbf{Task 2 } & \textbf{Task 3}     & \textbf{Task 4}   \\ \midrule 
\multicolumn{1}{l|}{Semantic split $\rightarrow$}      & \multicolumn{1}{c}{\begin{tabular}[c]{@{}c@{}}Animals, \\ Person, \\ Vehicles\end{tabular}} & \multicolumn{1}{c}{\begin{tabular}[c]{@{}c@{}}Appliances, \\ Accessories, \\ Outdoor, \\ Furniture\end{tabular}} & \multicolumn{1}{c}{\begin{tabular}[c]{@{}c@{}}Sports, \\ Food\end{tabular}} & \begin{tabular}[c]{@{}c@{}}Electronic, \\ Indoor, \\ Kitchen\end{tabular} \\ \midrule
\multicolumn{1}{l|}{$\#$ training images} & \multicolumn{1}{c}{89490}    & \multicolumn{1}{c}{55870}   & \multicolumn{1}{c}{39402}  & 38903    \\
\multicolumn{1}{l|}{$\#$ test images}     & \multicolumn{1}{c}{3793}    & \multicolumn{1}{c}{2351}   & \multicolumn{1}{c}{1642}    & 1691   \\
\multicolumn{1}{l|}{$\#$ train instances} & \multicolumn{1}{c}{421243}   & \multicolumn{1}{c}{163512}   & \multicolumn{1}{c}{114452}   & 160794      \\
\multicolumn{1}{l|}{$\#$ test instances}  & \multicolumn{1}{c}{17786}   & \multicolumn{1}{c}{7159}   & \multicolumn{1}{c}{4826}  & 7010   \\ \bottomrule 
\end{tabular}}
\caption{\textbf{The table shows task composition in the OWOD and MS-COCO split for the Open-world evaluation protocol}. The semantics of each task and the number of images and instances(objects) across splits are shown.} 
\label{table6}
\end{table}

\section{Experiment}
\subsection{Datasets and Metrics}
For a fair comparison, we implement the experiments on two mainstream splits of MS-COCO \cite{mscoco}, and Pascal VOC \cite{voc} dataset. We group the classes into a set of non-overlapping tasks $\left\{T^1, \ldots, T^t, \ldots\right\}$. The class in task $T^c$ only appears in tasks where $t \geq c$. In task $T^c$, classes encountered in $\left\{T^c: c \leq t\right\}$ and $\left\{T^c: c>t\right\}$ are considered as known and unknown classes, respectively. \par
\noindent \textbf{OWOD SPLIT} \cite{ORE} splits the 80 classes of MS-COCO into 4 tasks and selects a training set for each task from the MS-COCO and Pascal VOC training set. Pascal VOC testing and MS-COCO validation set are used for evaluation. Detailed data statistic is shown in TABLE.\ref{table6}\par
\noindent \textbf{MS-COCO SPLIT} \cite{owdetr} mitigates data leakage across tasks in \cite{ORE} and is more challenging, as shown in TABLE.\ref{table6}. The training and testing data are selected from MS-COCO.\par

\noindent \textbf{Metrics:} 
Following the most commonly used evaluation metric for object detection, we use mean average precision (mAP) to evaluate the known objects. Inspired by \cite{ORE,recall,owdetr,wi,A-OSE}, U-Recall which measures the ability of the model to retrieve unknown object instances for OWOD problems, is used as the metric for unknown objects. For the proposed \textbf{StandardSet}$\heartsuit$ and \textbf{IntensiveSet}$\spadesuit$ benchmark, we also use the unknown AP, unknown Precision to comprehensively measure the detection performance of open-world detectors for unknown open-world objects.
\subsection{Implementation Details}
The multi-scale feature extractor consists of a Resnet-50\cite{resnet50} pretrained on ImageNet\cite{imagenet} in a self-supervised\cite{self} manner and a deformable transformer encoder whose number of layers is set to 6. For the two cascade decoders, we all use the deformable transformer decoder, and the number of layers is set to 6, too. We set the number of queries $M=100$, the dimension of the embeddings $D=256$, and the number of pseudo-labels $k=5$. We set GLIP \cite{glip} as the large pre-trained vision-language grounding model and categories of LVIS dataset \cite{lvis} as text prompts to assist the training process. For GLIP, we use the GLIP-L \cite{glip} without finetuning on the COCO dataset \cite{mscoco}. It consists of Swin-Large \cite{swin}, text encoder of CLIP \cite{clip}, DyHead \cite{dai2021dynamic}, BERT Layer \cite{bert} and Fusion Module \cite{glip}. For the incremental object detection experiments, we only use our open-world detector without the help of GLIP. The PyTorch library and eight NVIDIA RTX 3090 GPUs are used to train our SKDF framework with a batch size of 3 images per GPU. In each task, the SKDF framework is trained for 50 epochs and finetuned for 20 epochs during the incremental learning step. We train our SKDF using the Adam optimizer with a base learning rate of $2\times10^{-4}$, $\beta_1=0.9$, $\beta_2=0.999$, and weight decay of $10^{-4}$. For finetuning during the incremental step, the learning rate is reduced by a factor of 10 and trained using a set of 50 stored exemplars per known class.

\begin{table*}[htbp]

\centering

\resizebox{0.8\textwidth}{!}{
\begin{tabular}{c|c| ccc|c|c|c }\toprule
\multicolumn{1}{c|}{}&Task IDs $\rightarrow$ &&&& \multicolumn{1}{c|}{Task 1}& \multicolumn{1}{c|}{Task 2} & \multicolumn{1}{c}{Task 3}   \\\midrule
\multicolumn{1}{c|}{}  &&&&Inference& \multicolumn{1}{c|}{Unknown}  & \multicolumn{1}{c|}{Unknown} &  \multicolumn{1}{c}{Unknown}    \\

 \multicolumn{1}{c|}{}& &\multicolumn{1}{c}{Param\#}  & \multicolumn{1}{c}{FLOPs} & \multicolumn{1}{c|}{Rate}  & Recall  &Recall&Recall    \\

\multicolumn{1}{c|}{\multirow{-3}{*}{SPLIT}}&\multirow{-3}{*}{Metrics $\rightarrow$} &&&(s/img) & \multirow{-1}{*}{($\uparrow$)} & \multirow{-1}{*}{($\uparrow$)}   & \multirow{-1}{*}{($\uparrow$)}  \\ \midrule
 \multirow{2}{*}{OWOD} & OV-VLM&321.9M&965GMac& 9.22&37.0    & 35.5    & 34.9 \\ 
 \multirow{1}{*}{}&\textbf{Ours}&  \textbf{42.9M} &\textbf{212GMac}& \textbf{0.08}    & \textbf{39.0}  &\textbf{36.7}    &\textbf{36.1} \\ \midrule
 \multirow{2}{*}{MS-COCO} & OV-VLM  &321.9M&965GMac& 9.22&52.6    & 54.5     & 53.3 \\ 
 \multirow{1}{*}{}&\textbf{Ours} &  \textbf{42.9M}&\textbf{212GMac}& \textbf{0.08} &\textbf{60.9}     &\textbf{60.0}    &\textbf{58.6} 
 \\\bottomrule
\end{tabular}}
\caption{\textbf{Comparison with the large pre-trained Open-Vocabulary grounding vision-language model for Open-World object detection on OWOD and MS-COCO split.} For the large pre-trained Open-Vocabulary grounding vision-language model, the prediction beyond the known categories is set to unknown. Experiments demonstrate that SKDF distills open-world knowledge from the large open-vocabulary pre-trained vision-language model to the expert open-world detector with faster-detecting speed and better performance for unknown open-world objects.} 
\label{table1}
\end{table*}

\begin{figure*}[htbp]
  \centering
  \includegraphics[width = \textwidth]{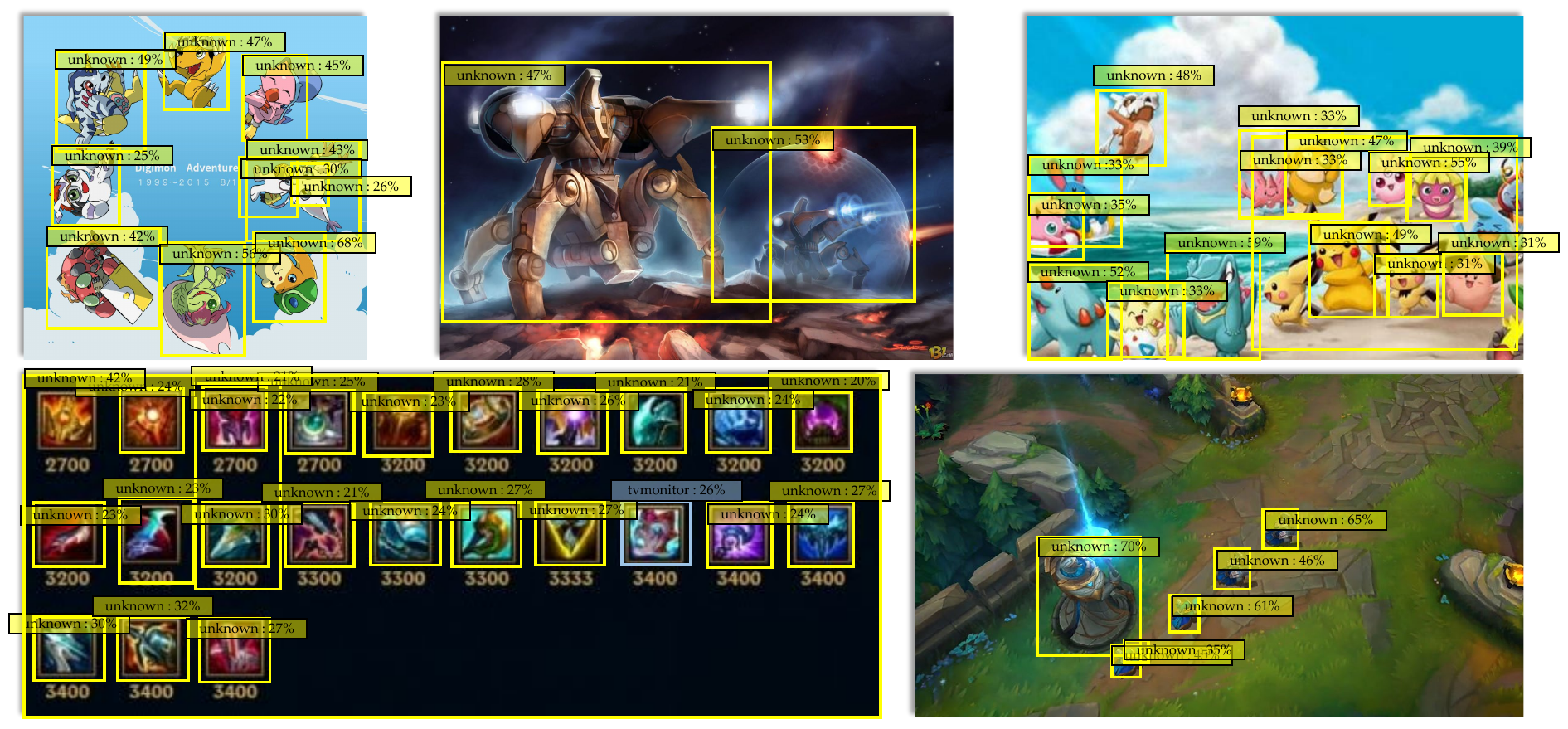}
  \caption{Qualitative Results. Visualization results are based on the setting of Task.1. Our model can detect the unknown objects in \textbf{Yellow} boxes beyond the unknown labels from GLIP and LVIS text prompts. The animation and games categories in the figures do not appear in the LVIS text prompt and our training dataset so our detector must not learn from GLIP.}
  \label{figure5}
\end{figure*}

\subsection{Detailed Comparison With the Teacher}
In this section, we validate the advantages of our distillation framework in terms of performance and speed through detailed comparison and quantitative experiments.
\subsubsection{Experimental Results}
In order to keep the model's ability to generalize to open-world objects in the comparison, we do not finetune the teacher on the known dataset. Because the finetuning process severely destroys the teacher's generality. Thus, we only shown the results on the comparison of the detection ability for unknown objects. The results compared with the large pre-trained vision-language grounding model (GLIP) for OWOD problem are shown in TABLE.\ref{table1}. Regarding the number of parameters and FLOPs, our model is significantly smaller than GLIP. In particular, the inference speed of ours is $115 \times \sim 116 \times$ of GLIP's. Furthermore, the number of GLIP's training data is 64M images, while our model only needs a small amount of data in each task of different splits almost $\frac{1}{237} \times \sim \frac{1}{16} \times$ of its.  Compared with GLIP's U-Recall of 37.0, 35.5, and 34.9 on Task 1, 2, and 3 of OWOD split, ours achieves 39.0, 36.7 and 36.1 in the corresponding tasks, achieving significant absolute gains up to 2.0, 1.2 and 1.2, respectively. In the MS-COCO split, compared with GLIP's U-Recall of 52.6, 54.5, and 53.3 on Task 1, 2, and 3 of OWOD split, ours achieves 60.9, 60.0 and 58.6 in the corresponding tasks, achieving significant absolute gains up to 8.3, 5.5 and 5.3, respectively. This demonstrates that our model has a better ability to detect unknown objects for OWOD.\par
\subsubsection{Qualitative Results}
To present the ability to detect unknown open-world objects intuitively, we select several open scenes from games and comics, such as ``Digital Monster'', ``Starcraft'', ``League of Legends'' and ``Pokemon''. These scenes contain the object out of the categories of LVIS \cite{lvis} text prompts, visualization results are shown in Fig.\ref{figure5}. The visualization results show that our detector can identify nearly all unknown open-world objects, such as the creatures from ``Digital Monster'' and ``Pokemon'', the robots from ``StarCraft'', and the minions and turrets from ``League of Legends''. Even the weapons and equipment from ``League of Legends'' can be detected by our detector; however, there are instances of misidentification, such as mistaking ``Liandry's Anguish'' for a TV monitor. Thus, qualitative results can demonstrate that SKDF evolves the novel unknown objects beyond the large pre-trained vision-language grounding model.

\subsection{Ablation Study}
In this subsection, a series of experiments are designed to clearly understand the contribution of each of the constituent components. We conducted all experiments on the OWOD split. We start by ablating each component, followed by ablating the text prompts and open-vocabulary large per-trained teacher. \par

\begin{table*}[htbp]
\centering
\resizebox{\textwidth}{!}{
\begin{tabular}{l| cc| cccc |cccc }\toprule
\multicolumn{1}{c|}{Task IDs $\rightarrow$}& \multicolumn{2}{c|}{Task 1}& \multicolumn{4}{c|}{Task 2} & \multicolumn{4}{c}{Task 3}  \\\midrule
\multicolumn{1}{c|}{}   & \multicolumn{1}{c}{ Unknown}  & Known &  \multicolumn{1}{c}{ Unknown} & \multicolumn{3}{c|}{Known} & \multicolumn{1}{c}{ Unknown} & \multicolumn{3}{c}{Known}    \\
\multicolumn{1}{l|}{}    &  Recall&  mAP($\uparrow$) & Recall& \multicolumn{3}{c|}{mAP($\uparrow$)}&Recall  &\multicolumn{3}{c}{mAP($\uparrow$)}   \\

\multicolumn{1}{c|}{\multirow{-3}{*}{Metrics $\rightarrow$}}   & \multirow{-1}{*}{($\uparrow$)}&Current & \multirow{-1}{*}{($\uparrow$)}   &Previously &Current &Both & \multirow{-1}{*}{($\uparrow$)}  &Previously & Current &Both \\ \midrule
\texttt{Baseline(GLIP)}      &  37.0        & 42.0    &  35.5 &42.1 & 22.6     &  32.4     & 34.9  &32.4  &19.2       & 28.2   \\ \midrule
\texttt{Distillation}      &  39.4         &41.1    &  36.6       &27.9&10.5 &  19.2     & 36.6       &18.0&6.2  &14.1   \\
\texttt{Distillation} \texttt{+} \texttt{DW}   &  39.2          &53.5   &    36.3        &47.5&22.3 &   34.9    &  35.9   &32.0&11.5    &25.1    \\
\texttt{Distillation} \texttt{+} \texttt{CS}    &39.1&51.9    &  36.9     &46.7&22.6  &   34.6    & 36.1        &32.4& 12.8      & 25.9    \\ 

\texttt{Final: SKDF}       &   39.0     &\textbf{56.8}    & 36.7       &\textbf{52.3} &\textbf{28.3}     &\textbf{40.3}    &   36.1   &\textbf{36.9}&\textbf{16.4}  &\textbf{30.1}      \\
\bottomrule
\end{tabular}}
\caption{\textbf{Experiments on ablating each component.} Our method significantly improves the detection performance of the GLIP baseline. \texttt{DW} represents the down-weight training loss function for unknown open-world supervision. When \texttt{DW} is none, we use the same loss function as the ground truth supervision for unknown open-world supervision. \texttt{CS} represents the cascade decoupling structure. When \texttt{CS} is none, we leverage the normal decoder structure as DDETR.} 
\label{table2}
\end{table*}

\begin{table*}[htbp]
\begin{subtable}{0.33\textwidth}
\centering
\resizebox{\textwidth}{!}{
\begin{tabular}{c | c c c c c }
\toprule
Prompt                                                                   & SPLIT                 & Method & mAP           & UR            & UR(\texttt{sam})       \\ \midrule \midrule
                                                                         &                       & GLIP   & 42.2          &  37.0          &  8.04          \\
                                                                         & \multirow{-2}{*}{\texttt{OWOD}} & Ours   & \textbf{56.8  } &  \textbf{39.0} &  \textbf{16.5} \\ 
                                                                         &                       & GLIP   & 46.5          &  52.6          &  9.4            \\
\multirow{-4}{*}{LVIS}                                                   & \multirow{-2}{*}{\texttt{MSCOCO}} & Ours   & \textbf{69.4} &  \textbf{60.9} &  \textbf{20.0} \\ \midrule
                                                                         &                       & GLIP   & -             &  -             &  7.5             \\
                                                                         & \multirow{-2}{*}{\texttt{OWOD}} & Ours   & \textbf{57.6} &  \textbf{38.8} &  \textbf{16.8} \\ 
                                                                         &                       & GLIP   & -             &  -             &  7.6             \\
\multirow{-4}{*}{\begin{tabular}[c]{@{}c@{}}LVIS\\ \texttt{w/o} COCO\end{tabular}} & \multirow{-2}{*}{\texttt{MSCOCO}} & Ours   & \textbf{70.6} &  \textbf{58.9} &  \textbf{20.4} \\ \midrule
                                                                         &                       & GLIP   & \textbf{67.8} &  \textbf{43.3} &  3.2           \\
                                                                         & \multirow{-2}{*}{\texttt{OWOD}} & Ours   & 60.9          &  41.1          &  \textbf{12.9} \\  
                                                                         &                       & GLIP   & 73.1          &  64.8          &  5.3             \\
\multirow{-4}{*}{COCO}  & \multirow{-2}{*}{\texttt{MSCOCO}} & Ours   & \textbf{74.8} &  \textbf{69.6} &  \textbf{16.7} \\ \bottomrule
\end{tabular}}
\caption{\textbf{Ablating different prompts}.}
\end{subtable}%
\hfill
\begin{subtable}{0.3\textwidth}
\centering
\resizebox{\textwidth}{!}{
\begin{tabular}{c|cccc}
\toprule
\multicolumn{1}{l|}{SPLIT}      &   Metrics    &\texttt{DDETR}  &\texttt{CAT}                                 & Ours                             \\ \midrule \midrule
                            & mAP     & 53.5  & 55.3          &  \textbf{56.8}  \\
                            &UR       & 39.2 & \textbf{39.4} &39.0          \\
\multirow{-3}{*}{\texttt{OWOD}} & UR(\texttt{sam}) & 16.3 & 16.5          &\textbf{16.5} \\ \bottomrule
\end{tabular}}
\caption{\textbf{Ablating different structures}.}
\resizebox{\textwidth}{!}{
\begin{tabular}{c|cccc}
\toprule
                        &                         & \multicolumn{3}{c}{Metric}                                                                                            \\
\multirow{-2}{*}{\texttt{Teacher}} & \multirow{-2}{*}{SPLIT} &mAP           & UR            & UR(sam)       \\ \midrule \midrule
GLIP                    & \texttt{OWOD}                    & \textbf{56.8} & \textbf{39.0} &16.5          \\
SAM                     & \texttt{OWOD}                    & 19.4          &33.5          &\textbf{38.3} \\ \bottomrule
\end{tabular}}
\caption{\textbf{Ablating different Teacher}.}
\end{subtable}
\hfill
\begin{subtable}{0.3\textwidth}
    \centering
    \includegraphics[width=\textwidth]{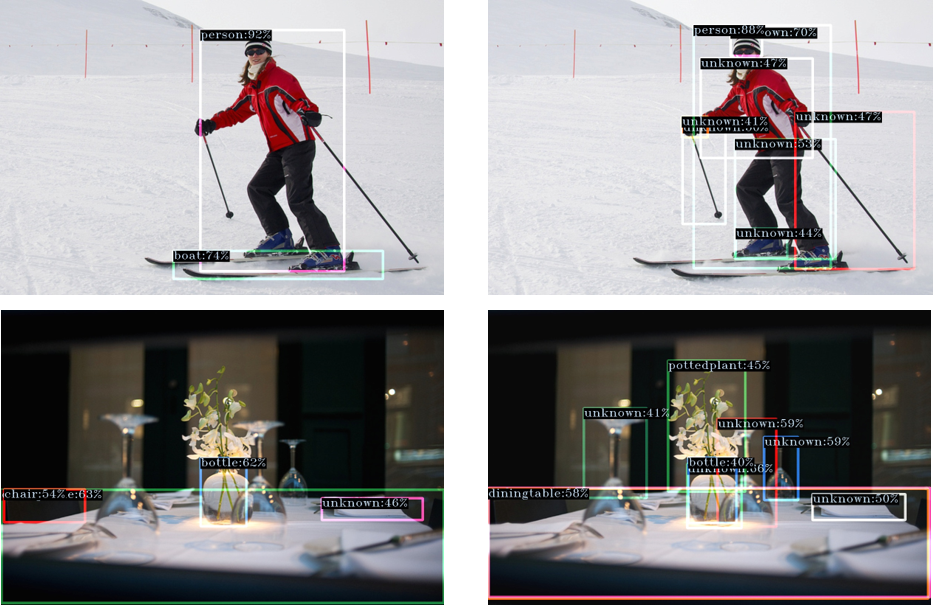}
    \caption{Left(Right) learns from \texttt{COCO(LVIS)} prompt.}
\end{subtable}
\caption{\textbf{Detailed ablation experiments for our SKDF.} \texttt{(a)} Ablation experiments on different text prompts, where UR denotes the unknown recall on original unknown annotations, UR(sam) denotes the unknown recall on unknown annotations with SAM generation. \texttt{(b)} Ablation experiments on different detector structures. \texttt{(c)} Ablation experiments on different teacher models. \texttt{(d)} Qualitative ablation on \texttt{LVIS} and \texttt{COCO} prompt.} \label{table3}
\end{table*}

\subsubsection{Ablating components}
To study the contribution of each component, we design ablation experiments in TABLE.\ref{table2}. We set GLIP as our baseline, the \texttt{Distillation} improves the performance on unknown object detection but reduces the detector's ability for known objects. Adding the down-weight training loss function significantly improves the performance on detecting known objects and incremental object detecting, achieving significant absolute gains up to more than 10 points. As we have analyzed, the cascade decoupling structure alleviates the inclusion of unknown objects on the known detecting performance and the confusion between the categories and locations of the same objects. It significantly improves the performance of detecting known objects with absolute gains up to more than 10 points, too. What's more, these two combine effectively, improving performance with absolute gains up to 15.7, 21.1, and 16.0, without reducing the ability to detect unknown objects. Thus, each component has a critical role to play in open-world object detection. \par

\begin{table*}[htbp]
\renewcommand\arraystretch{1.25}
\centering
\resizebox{\textwidth}{!}{
\begin{tabular}{l| cc| cccc |cccc |ccc}\toprule
\multicolumn{1}{c|}{Task IDs $\rightarrow$}& \multicolumn{2}{c|}{Task 1}& \multicolumn{4}{c|}{Task 2} & \multicolumn{4}{c|}{Task 3}  & \multicolumn{3}{c}{Task 4} \\\midrule
\multicolumn{1}{c|}{}   & \multicolumn{1}{c}{ Unknown}  & Known &  \multicolumn{1}{c}{ Unknown} & \multicolumn{3}{c|}{Known} & \multicolumn{1}{c}{ Unknown} & \multicolumn{3}{c|}{Known}  &\multicolumn{3}{c}{Known}  \\
\multicolumn{1}{l|}{}    &  Recall&  mAP($\uparrow$) & Recall& \multicolumn{3}{c|}{mAP($\uparrow$)}&Recall  &\multicolumn{3}{c|}{mAP($\uparrow$)}   &  \multicolumn{3}{c}{mAP($\uparrow$)}\\

\multicolumn{1}{c|}{\multirow{-3}{*}{Metrics $\rightarrow$}}   & \multirow{-1}{*}{($\uparrow$)}& Current & \multirow{-1}{*}{($\uparrow$)}   &Previously & Current &Both & \multirow{-1}{*}{($\uparrow$)}  &Previously & Current &Both &Previously & Current &Both \\ \midrule

UC-OWOD\cite{UC-OWOD} &2.4  & 50.7 & 3.4  & 33.1& 30.5 & 31.8 &8.7  &28.8&16.3&24.6&25.6&15.9&23.2 \\
ORE-EBUI\cite{ORE}& 4.9  & 56.0   &2.9  & 52.7 & 26.0  & 39.4  & 3.9   & 38.2  & 12.7 & 29.7  & 29.6   & 12.4  & 25.3  \\
OW-DETR\cite{owdetr}  &7.5 &  59.2 & 6.2 & 53.6  & 33.5   & 42.9  & 5.7  & 38.3 & 15.8 & 30.8  & 31.4 & 17.1 & 27.8  \\ 
OCPL\cite{OCPL} &8.3  & 56.6 &7.7  & 50.6    & 27.5 & 39.1 &11.9  &38.7&14.7&30.7&30.7&14.4 &26.7\\
2B-OCD\cite{two-branch} &12.1  & 56.4 & 9.4  & 51.6& 25.3 & 38.5 &11.6  & 37.2 & 13.2 & 29.2&30.0&13.3&25.8 \\
 \multirow{1}{*}{\textbf{Ours}}   &\textbf{39.0}    &56.8 & \textbf{36.7} & 52.3 & 28.3  & 40.3  & \textbf{36.1}&  36.9 & 16.4  & 30.1 &31.0 & 14.7  &26.9\\ 

\bottomrule 
\end{tabular}}
\caption{\textbf{State-of-the-art comparison for open-world object detection on OWOD split.} The comparison is shown in terms of U-Recall and known class mAP. U-Recall measures the ability of the model to retrieve unknown object instances for OWOD problems. For a fair comparison, we compare with the recently introduced methods. The CAT achieves improved metrics over the existing works across all unknown detection tasks, demonstrating our model’s effectiveness for OWOD problems. U-Recall cannot be computed in Task 4 due to the absence of unknown test annotations, for the reason that all 80 classes are known.} 
\label{table4}
\end{table*}

\begin{table*}[h]
\renewcommand\arraystretch{1.25}
\centering
\resizebox{\textwidth}{!}{
\begin{tabular}{l| cc| cccc |cccc |ccc}\toprule
\multicolumn{1}{c|}{Task IDs $\rightarrow$}& \multicolumn{2}{c|}{Task 1}& \multicolumn{4}{c|}{Task 2} & \multicolumn{4}{c|}{Task 3}  & \multicolumn{3}{c}{Task 4} \\\midrule
\multicolumn{1}{c|}{}   & \multicolumn{1}{c}{ Unknown}  & Known &  \multicolumn{1}{c}{ Unknown} & \multicolumn{3}{c|}{Known} & \multicolumn{1}{c}{ Unknown} & \multicolumn{3}{c|}{Known}  &\multicolumn{3}{c}{Known}  \\
\multicolumn{1}{l|}{}    &  Recall&  mAP($\uparrow$) & Recall& \multicolumn{3}{c|}{mAP($\uparrow$)}&Recall  &\multicolumn{3}{c|}{mAP($\uparrow$)}   &  \multicolumn{3}{c}{mAP($\uparrow$)}\\

\multicolumn{1}{c|}{\multirow{-3}{*}{Metrics $\rightarrow$}}   & \multirow{-1}{*}{($\uparrow$)}& Current & \multirow{-1}{*}{($\uparrow$)}   &Previously & Current &Both & \multirow{-1}{*}{($\uparrow$)}  &Previously & Current &Both &Previously & Current &Both \\ \midrule

ORE-EBUI\cite{ORE}& 1.5 & 61.4   & 3.9& 56.5 & 26.1  & 40.6  &3.6   & 38.7  & 23.7 & 33.7   & 33.6   & 26.3  & 31.8  \\
OW-DETR\cite{owdetr}  & 5.7 &  71.5 & 6.2 & 62.8  & 27.5   & 43.8  & 6.9& 45.2 & 24.9 & 38.5  & 38.2 & 28.1 & 33.1  \\ 

 \multirow{1}{*}{\textbf{Ours}}   &\textbf{60.9}  &69.4 & \textbf{60.0} &63.8 & 26.9  & 44.4  & \textbf{58.6}&  46.2 &28.0 & 40.1 &41.8 &29.6  &38.7\\ 
\bottomrule 
\end{tabular}}
\caption{\textbf{State-of-the-art comparison for open-world object detection on MS-COCO split.} As the code and weights of UC-OWOD \cite{UC-OWOD}, OCPL \cite{OCPL} and 2B-OCD \cite{two-branch} are not publicly available, we cannot get results of them or evaluate them on MS-COCO split. Thus, we only compare our model with ORE \cite{ORE} and OW-DETR \cite{owdetr}. Although the MS-COCO split is more challenging, our model gets a more significant improvement on this in comparison to ORE and OW-DETR. The significant metric improvements demonstrate that our SKDF has the ability to retrieve new knowledge beyond the range of closed sets and would not be limited by category knowledge of existing objects.} 
\label{table7}

\end{table*}

\subsubsection{Ablating different prompts}
To ablate the impact of text prompts for knowledge distillation, we do experiments on different text prompts, \ie LVIS, LVIS without COCO, and COCO as shown in TABLE.\ref{table3}\texttt{(a)}. From the experiment results (\ie UR comparison between LVIS and COCO prompt and qualitative results in TABLE.\ref{table3}\texttt{(d)}), we investigate that the unknown annotations in the test set are not adequate and more corresponding to the original 80 categories in COCO dataset. Therefore, existing ``unknown recall'' can not evaluate the detecting ability for unknown open-world objects beyond COCO annotation. To evaluate that, we use SAM\cite{SAM} to generate annotations for all areas of similar objects in the test set. Though these annotations are noisy and contain many non-object box annotations, the recall for these can evaluate the unknown detection ability of detectors. The comparison between LVIS and COCO prompts shows that less distilled knowledge reduces the impact on learning known objects. In addition, less distilled knowledge impacts the performance of the knowledge distillation framework. As shown in \texttt{OWOD} split of \texttt{COCO} prompt, our detector does not learn better detecting performance for the original unknown objects. The results in \texttt{LVIS w/o COCO} prompt demonstrate our framework letter more unknown open-world objects. Although we exclude all corresponding coco categories from the LVIS prompt, our detector's detection ability for original unknown objects does not seem to be impacted.

\subsubsection{Ablating different detection transformer structures}
As shown in TABLE.\ref{table3}\texttt{(b)}, we compare our structure with original parallel structure \cite{ddetr} and structure of \cite{CAT}. CAT proposes the cascade decoupled decoding way that uses the shared decoder to decode twice for localization and identification, respectively. Unlike them, we directly employ two separate decoders to decouple the decoding process, and we specifically train these two decoders for localization and recognition, respectively. The results demonstrate that our structure has a better ability to alleviate the influence of distilled knowledge on known objects. Our decoupled architecture outperforms the decoupled decoding way of CAT\cite{CAT} by 1.5 points in the known mAP. Moreover, our architecture surpasses the conventional DDETR \cite{ddetr} structure by 3.3 points.

\subsubsection{Ablating different teachers}
In TABLE.\ref{table3}\texttt{(c)}, we compare different teachers, \ie GLIP\cite{glip} and SAM \cite{SAM}. When distilling open-world knowledge from SAM, we leverage the ``Segment Anything'' that uses the predefined grid points and selecting strategy to generate pseudo labels. For SAM, due to the absence of supervision confidence, we can not leverage the down-weight loss function. Therefore, the known detection ability is severely reduced. In addition, due to the over-detection ability of SAM, the detector can not distill a better unknown detection ability.

\subsection{Comparison With SOTA OWOD Methods}

In this section, we conduct a detailed comparison with existing state-of-the-art methods on two existing benchmarks: OWOD and MS-COCO SPLIT, as well as on two benchmarks we propose: \textbf{IntensiveSet}$\spadesuit$ and \textbf{StandardSet}$\heartsuit$. In addition, we compare our model with existing methods on the Incremental Object Detection task.

\subsubsection{OWOD SPLIT}
The results compared with the state-of-the-art methods on OWOD split for OWOD problem are shown in TABLE.\ref{table4}. The ability of our model to detect unknown objects quantified by U-Recall is more than $3 \times$ of those reported in previous state-of-the-art OWOD methods.  Compared with the model 2B-OCD \cite{two-branch} with the highest U-Recall of 12.1, 9.4 and 11.6 on Task 1, 2, and 3, ours achieves 39.0, 36.7, and 36.1 in the corresponding tasks, achieving significant absolute gains up to 26.9, 27.3 and 24.5, respectively. Benefits from the cascade decoder structure and down-weight training loss which mitigate the effect of unknown objects on detecting known objects, our model's performance on known objects is also superior to most state-of-the-art methods. 
\par

\subsubsection{MS-COCO SPLIT}
We report the results on the MS-COCO split in TABLE.\ref{table7}. MS-COCO split mitigates data leakage across tasks and assigns more data to each Task, while our model receives a more significant boost compared with the OWOD split. Our model's unknown object detection capability, quantified by U-Recall, is almost $10 \times \sim 11 \times$ of those reported in previous state-of-the-art OWOD methods. Compared with OW-DETR's U-Recall of 5.7, 6.2, and 6.9 on Task 1, 2, and 3, ours achieves 60.9, 60.0, and 58.6 in the corresponding tasks, achieving significant absolute gains up to 55.2, 53.8 and 51.7, respectively. This demonstrates that our model has the more powerful ability to retrieve new knowledge and detect unknown objects when faced with more difficult tasks.
\par

\begin{table*}[htbp]
\centering
\resizebox{0.9\textwidth}{!}{
\begin{tabular}{c|c|cccc|cccc}
\toprule
 & Dataset$\rightarrow$                  & \multicolumn{4}{c|}{\texttt{IntensiveSet}$\spadesuit$}     & \multicolumn{4}{c}{\texttt{StandardSet}$\heartsuit$}      \\ \midrule
\multirow{2}{*}{\begin{tabular}[c]{@{}c@{}}Group\\ $\downarrow$\end{tabular}} &\multirow{2}{*}{Metrics$\rightarrow$} & Known  & \multicolumn{3}{c|}{Unknown($\uparrow$)} & Known & \multicolumn{3}{c}{Unknown($\uparrow$)}        \\
 & & mAP($\uparrow$)   & Recall   & Precision  & mAP  & mAP($\uparrow$)   & Recall   & Precision  & mAP     \\
\midrule
\multirow{4}{*}{(\texttt{\uppercase\expandafter{\romannumeral1}})}&OWDETR\cite{owdetr}                   &    36.9     &       5.6  &        7.3    &    1.2  &    54.6   & 16.9         &   2.3         &  0.6      \\
&CAT\cite{CAT}                      &  38.6   &  17.4   &   9.5         &  7.6   & \textbf{55.4} &   48.8    &   2.7       & 4.3                   \\
&UnSniffer\cite{unsniffer}                &    \textbf{39.5}    &    9.4      &     22.2       &   2.0   &    53.7   &      34.6    &   \textbf{21.5}         &   12.5       \\ 
&UnSniffer$\dagger$\cite{unsniffer}               &    39.5    &    12.4      &     12.0       &   2.6   &    53.7   &      41.2    &   9.6         &   11.6      \\ \midrule
\multirow{3}{*}{(\texttt{\uppercase\expandafter{\romannumeral2}})}&SKDF-DW                &    27.7    & \textbf{40.2}         &     15.6       &   14.8   &  44.0     &     75.1     &    3.0        &   17.1          \\ 
&SKDF-CS                &    30.7    &     39.7     &      16.2      &  16.8    &    48.3   &      \textbf{75.2}    &      3.1      &   9.2     \\ 
&SKDF                     &   32.3     &     39.6     &    16.2        &   16.7   &   48.7    &    74.3      &      3.0      &   7.4      \\ 
&SKDF$\ddagger$                   &   32.3     &    35.8     &    \textbf{37.9}        &   \textbf{24.5}   &   48.7    &    66.8      &     11.4     &   \textbf{24.4}      \\
\bottomrule
\end{tabular}}
\caption{\textbf{State-of-the-art comparison for open-world object detection on \texttt{IntensiveSet}$\spadesuit$ and \texttt{StandardSet}$\heartsuit$.} As the code and weights of UC-OWOD \cite{UC-OWOD}, OCPL \cite{OCPL}, and 2B-OCD \cite{two-branch} are not publicly available, we cannot get results of them or evaluate them on our proposed benchmark. For the fair comparison, UnSniffer$\dagger$ means that we remove the specific unknown post-process operations. SKDF$\ddagger$ denotes that we add the same unknown post-process operations as Unisiffer to our SKDF.} 
\label{table8}
\end{table*}

\subsubsection{Intensive Testing Set}
The comparison results on \textbf{IntensiveSet}$\spadesuit$ are shown in the left half of TABLE.\ref{table8}. The experimental results indicate that our model's performance in detecting unknown objects surpasses that of existing state-of-the-art methods, achieving more than double the performance of the highest unknown recall for the category CAT \cite{CAT}, exceeding the highest unknown precision of Unsiffer \cite{unsniffer} by 4.2 points, and surpassing the existing methods in unknown mAP by more than threefold. Meanwhile, our model is highly adaptable to unknown post-processing techniques. When incorporating the post-processing methods from UnSniffer \cite{unsniffer}, the performance of our model further increased by 21.7 points in unknown precision and 7.8 points in unknown mAP. When incorporating the post-processing, SKDF surpasses the UnSiffer in unknown precision by more than 15 points. However, although our component can mitigate the impact of unknown objects on the detection of known objects, there is still a certain gap between our model and the existing state-of-the-art models.

\subsubsection{Standard Testing Set}
In the right half of the TABLE.\ref{table8}, we compare SKDF with the sota methods on \textbf{StandardSet}$\heartsuit$. Our model surpasses the existing detection models in the vast majority of evaluation metrics, except for the detection performance of known objects and unknown precision. For the unknown precision, UnSiffer outperforms our SKDF. This is because the \textbf{StandardSet}$\heartsuit$ primarily comprises common categories from COCO that are beyond the predefined classes of VOC, and UnSiffer is specifically tailored to work with these category distributions. However, this indeed reflects the shortcomings of our model. Particularly, when comparing the performance of our model on the \textbf{IntensiveSet}$\spadesuit$ and \textbf{StandardSet}$\heartsuit$, it is evident that although our model has very good recall performance for unknown objects, its ability to recognize unknown objects still needs to be improved.

\subsubsection{Incremental Object Detection}
To intuitively present our detector's ability for detecting object instances, we compare it to \cite{iod,fasteriod,ORE,owdetr} on the incremental object detection (IOD) task. We do not use assistance from the large pre-trained vision-language grounding model. We evaluate the experiments on three standard settings, where a group of classes (10, 5, and last class) is introduced incrementally to a detector trained on the remaining classes (10, 15, and 19), based on the PASCAL VOC 2007 dataset \cite{voc}. As the results shown in TABLE.\ref{table9} (b), our model outperforms the existing method in a great migration on all three settings, indicating the power of cascade detection transformer for IOD.\par

\begin{table*}[htbp]
\renewcommand\arraystretch{1.25}

\resizebox{\textwidth}{!}{
\begin{tabular}{lccccccccccccccccccccc}
\toprule
\large{\textcolor{deepgreen}{\textbf{10 + 10 Setting}}}  & aero & cycle & bird & boat & bottle & bus  & car  & cat  & chair & cow  & table  & dog  & horse  & bike   & person & plant  & sheep  & sofa   & train  & tv & mAP  \\
\midrule
ILOD & 69.9 & 70.4  & 69.4 & 54.3 & 48   & 68.7 & 78.9 & 68.4 & 45.5  & 58.1 & \cellcolor[HTML]{EFEFEF}59.7&\cellcolor[HTML]{EFEFEF}72.7 &\cellcolor[HTML]{EFEFEF}73.5 &\cellcolor[HTML]{EFEFEF}73.2 &\cellcolor[HTML]{EFEFEF}66.3 &\cellcolor[HTML]{EFEFEF}29.5 &\cellcolor[HTML]{EFEFEF}63.4 &\cellcolor[HTML]{EFEFEF}61.6 &\cellcolor[HTML]{EFEFEF}69.3 &\cellcolor[HTML]{EFEFEF}62.2 & 63.2 \\
Faster ILOD  & 72.8 & 75.7  &  71.2  &  60.5  &  61.7  &  70.4  &  83.3  &  76.6  &  53.1  &  72.3  &\cellcolor[HTML]{EFEFEF} 36.7  &\cellcolor[HTML]{EFEFEF} 70.9  &\cellcolor[HTML]{EFEFEF} 66.8  &\cellcolor[HTML]{EFEFEF} 67.6  &\cellcolor[HTML]{EFEFEF} 66.1  &\cellcolor[HTML]{EFEFEF}24.7  &\cellcolor[HTML]{EFEFEF} 63.1  &\cellcolor[HTML]{EFEFEF} 48.1  &\cellcolor[HTML]{EFEFEF} 57.1  &\cellcolor[HTML]{EFEFEF} 43.6  & 62.1 \\
ORE - (CC + EBUI)  &53.3& 69.2& 62.4& 51.8& 52.9& 73.6& 83.7& 71.7& 42.8& 66.8& \cellcolor[HTML]{EFEFEF}46.8&\cellcolor[HTML]{EFEFEF}59.9&\cellcolor[HTML]{EFEFEF}65.5&\cellcolor[HTML]{EFEFEF}66.1&\cellcolor[HTML]{EFEFEF}68.6&\cellcolor[HTML]{EFEFEF}29.8&\cellcolor[HTML]{EFEFEF}55.1&\cellcolor[HTML]{EFEFEF}51.6&\cellcolor[HTML]{EFEFEF}65.3&\cellcolor[HTML]{EFEFEF}51.5& 59.4  \\
ORE - EBUI &63.5& 70.9& 58.9& 42.9& 34.1& 76.2& 80.7& 76.3& 34.1& 66.1&\cellcolor[HTML]{EFEFEF}56.1&\cellcolor[HTML]{EFEFEF}70.4&\cellcolor[HTML]{EFEFEF}80.2&\cellcolor[HTML]{EFEFEF}72.3&\cellcolor[HTML]{EFEFEF}81.8&\cellcolor[HTML]{EFEFEF}42.7&\cellcolor[HTML]{EFEFEF}71.6&\cellcolor[HTML]{EFEFEF}68.1&\cellcolor[HTML]{EFEFEF}77&\cellcolor[HTML]{EFEFEF}67.7& 64.5  \\

OW - DETR &75.4&63.9&57.9&50.0&52.0&70.9&79.5&72.4&44.3&57.9&\cellcolor[HTML]{EFEFEF}59.7&\cellcolor[HTML]{EFEFEF}73.5&\cellcolor[HTML]{EFEFEF}77.7&\cellcolor[HTML]{EFEFEF}75.2&\cellcolor[HTML]{EFEFEF}76.2&\cellcolor[HTML]{EFEFEF}44.9&\cellcolor[HTML]{EFEFEF}68.8&\cellcolor[HTML]{EFEFEF}65.4&\cellcolor[HTML]{EFEFEF}79.3&\cellcolor[HTML]{EFEFEF}69.0   &65.7  \\\midrule
\textbf{Ours} & 77.1& 72.3&74.5&53.4&57.4& 78.1& 78.7 &83.9& 46.2 & 71.4  &\cellcolor[HTML]{EFEFEF}59.5  &\cellcolor[HTML]{EFEFEF}77.4   &\cellcolor[HTML]{EFEFEF}73.3   &\cellcolor[HTML]{EFEFEF}76.6   &\cellcolor[HTML]{EFEFEF}73.3   &\cellcolor[HTML]{EFEFEF}39.7   &\cellcolor[HTML]{EFEFEF}70.6   &\cellcolor[HTML]{EFEFEF}59.0 &\cellcolor[HTML]{EFEFEF}78.4   &\cellcolor[HTML]{EFEFEF}70.9   &68.6 \\\midrule\midrule

\large{\textcolor{deepgreen}{\textbf{15 + 5 Setting}}}& aero & cycle & bird & boat & bottle & bus  & car  & cat  & chair & cow  & table  & dog  & horse  & bike & person & plant  & sheep  & sofa & train  & tv & mAP  \\\midrule
ILOD & 70.5 & 79.2 &68.8& 59.1& 53.2& 75.4& 79.4& 78.8& 46.6& 59.4& 59 &75.8& 71.8& 78.6& 69.6&\cellcolor[HTML]{EFEFEF}33.7&\cellcolor[HTML]{EFEFEF}61.5&\cellcolor[HTML]{EFEFEF}63.1 &\cellcolor[HTML]{EFEFEF}71.7&\cellcolor[HTML]{EFEFEF}62.2& 65.8 \\
Faster ILOD  &66.5& 78.1& 71.8& 54.6& 61.4& 68.4& 82.6& 82.7& 52.1& 74.3& 63.1& 78.6& 80.5& 78.4& 80.4&\cellcolor[HTML]{EFEFEF}36.7&\cellcolor[HTML]{EFEFEF}61.7&\cellcolor[HTML]{EFEFEF}59.3&\cellcolor[HTML]{EFEFEF}67.9&\cellcolor[HTML]{EFEFEF}59.1 &67.9 \\
ORE - (CC + EBUI)  &65.1& 74.6& 57.9& 39.5 &36.7& 75.1& 80 &73.3& 37.1& 69.8& 48.8& 69 &77.5& 72.8& 76.5&\cellcolor[HTML]{EFEFEF}34.4&\cellcolor[HTML]{EFEFEF}62.6&\cellcolor[HTML]{EFEFEF}56.5&\cellcolor[HTML]{EFEFEF}80.3&\cellcolor[HTML]{EFEFEF}65.7& 62.6 \\
ORE - EBUI &75.4& 81& 67.1& 51.9& 55.7& 77.2& 85.6& 81.7& 46.1& 76.2& 55.4& 76.7& 86.2& 78.5& 82.1&\cellcolor[HTML]{EFEFEF}32.8&\cellcolor[HTML]{EFEFEF}63.6&\cellcolor[HTML]{EFEFEF}54.7&\cellcolor[HTML]{EFEFEF}77.7&\cellcolor[HTML]{EFEFEF}64.6 &68.5 \\
OW - DETR &78.0&80.7&79.4&70.4&58.8&65.1&84.0&86.2&56.5&76.7&62.4&84.8&85.0&81.8&81.0&\cellcolor[HTML]{EFEFEF}34.3&\cellcolor[HTML]{EFEFEF}48.2&\cellcolor[HTML]{EFEFEF}57.9&\cellcolor[HTML]{EFEFEF}62.0&\cellcolor[HTML]{EFEFEF}57.0 &69.4\\\midrule
\textbf{Ours} &79.5&85.1&83.1 &73.1 &62.5 &68.7 &83.0 &88.4 &55.5 &78.3&69.7 &83.0 &86.6 &73.2&78.8 &\cellcolor[HTML]{EFEFEF}30.8&\cellcolor[HTML]{EFEFEF}67.6&\cellcolor[HTML]{EFEFEF}60.8&\cellcolor[HTML]{EFEFEF}76.0&\cellcolor[HTML]{EFEFEF}58.7 &72.1 \\\midrule\midrule

\large{\textcolor{deepgreen}{\textbf{19 + 1 Setting}}} & aero & cycle & bird & boat & bottle & bus  & car  & cat  & chair & cow  & table  & dog  & horse  & bike   & person & plant  & sheep  & sofa   & train  & tv & mAP  \\\midrule
ILOD & 69.4&79.3&69.5&57.4&45.4&78.4&79.1&80.5&45.7&76.3&64.8&77.2&80.8&77.5&70.1&42.3&67.5&64.4&76.7&\cellcolor[HTML]{EFEFEF}62.7&68.2  \\
Faster ILOD   &64.2&74.7&73.2&55.5&53.7&70.8&82.9&82.6&51.6&79.7&58.7&78.8&81.8&75.3&77.4&43.1&73.8&61.7&69.8&\cellcolor[HTML]{EFEFEF}61.1&68.5  \\
ORE - (CC + EBUI)  &60.7&78.6&61.8&45&43.2&75.1&82.5&75.5&42.4&75.1&56.7&72.9&80.8&75.4&77.7&37.8&72.3&64.5&70.7&\cellcolor[HTML]{EFEFEF}49.9&64.9 \\
ORE - EBUI   &67.3&76.8&60&48.4&58.8&81.1&86.5&75.8&41.5&79.6&54.6&72.8&85.9&81.7&82.4&44.8&75.8&68.2&75.7&\cellcolor[HTML]{EFEFEF}60.1&68.8\\

OW - DETR &82.2&80.7&73.9&56.0&58.6&72.1&82.4&79.6&48.0&72.8&64.2&83.3&83.1&82.3&78.6&42.1&65.5&55.4&82.9&\cellcolor[HTML]{EFEFEF}60.1  &70.2  \\\midrule

\textbf{Ours} &83.6 &85.7 &77.1 &61.5&58.9 &74.3
&86.3 &81.5 &52.2 &78.4 &71.4 &81.9 &84.6&80.2 &80.8 &39.9 &68.3&63.3 &84.6&\cellcolor[HTML]{EFEFEF}63.0  &72.9  \\ \bottomrule
\end{tabular}
}
\caption{\textbf{The detailed comparison with existing approaches on PASCAL VOC. We only use our open-world object detector without assistance.}  Evaluation is performed on three standard settings, where a group of classes (10, 5, and last class) are introduced incrementally to a detector trained on the remaining classes (10,15 and 19). Our model performs favorably against existing approaches on all three settings.} 
\label{table9}
\end{table*}

\subsubsection{Qualitative Comparison with OWOD method}
\begin{figure*}[htbp]
  \centering
  \includegraphics[width = \textwidth]{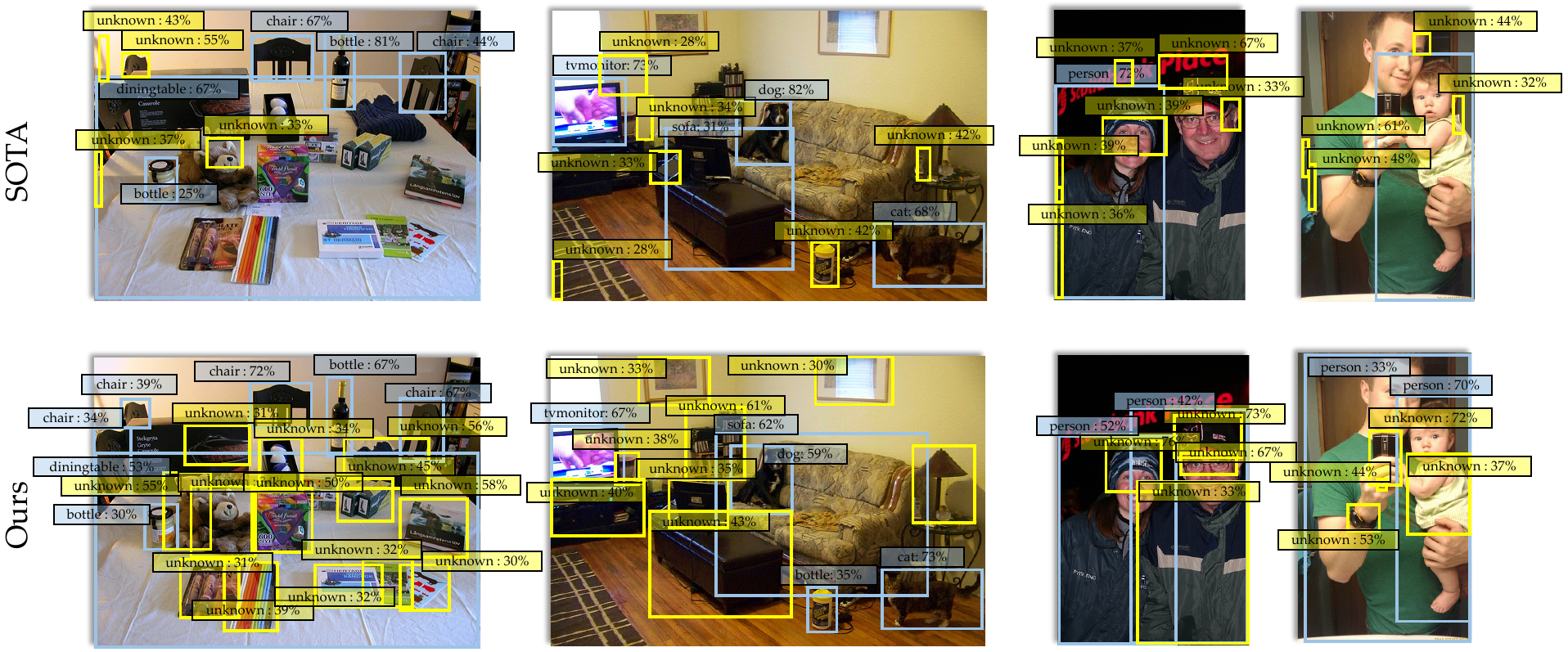}
  \caption{\textbf{Visualization results for the comparison between the SOTA method and our model} (known objects in \textbf{Blue} and unknown objects in \textbf{Yellow}). The categories of PASCAL VOC \cite{voc} are set as known.  Our model significantly outperforms the SOTA (OW-DETR) for open-world object detection, almost accurately detecting all unknown objects in the open scene, including \textit{clothes} and \textit{hats} worn by people, while most of the unknown objects detected by the SOTA model are backgrounds.}
  \label{figure6}
\end{figure*}

We exhibit more qualitative results in Fig.\ref{figure6} and compare them to the state-of-the-art OWOD detectors. The results show that our model can detect most of the potentially unknown open-world objects in the scene, far exceeding the effectiveness of existing models. We define the 20 categories of the PASCAL VOC \cite{voc} dataset as known classes, and all categories outside of these are considered unknown classes. As shown in the first column, our model accurately detects all the foreground categories in the scene and identifies everything outside of the predefined known classes as unknown, such as the books on the table and the teddy bear, which were not detected by the state-of-the-art model. As shown in the second column, our model can accurately detect even objects such as the picture frames on the wall. In the third column, the state-of-the-art model mistakenly identifies the characters in the background as unknown objects, but our model is not confused by this. In the comparison of the last column, our model also greatly surpasses the detection performance of the state-of-the-art model.
 
\section{Societal Impact, Limitations and Future Works}
Open-world object detection makes artificial intelligence smarter to face more problems in real life. It takes object detection to a cognitive level, as the model requires more than simply remembering the objects learned, it requires deeper thinking about the scene. When it comes to applications like autonomous driving, the significance of open-world object detection becomes even more pronounced. In such scenarios, vehicles need to rapidly and accurately identify and comprehend various objects and obstacles on the road, including but not limited to pedestrians, other vehicles, traffic signals, and road signs. The breakthrough in open-world object detection will render autonomous driving systems more intelligent as they can handle unforeseen or rare situations, not limited to pre-trained object categories.\par
Although our results demonstrate significant improvements over existing state-of-the-art methods, the performances are still on the lower side due to the challenging nature of the open-world detection problem. In this paper, we are mainly committed to enhancing the model's ability to explore unknown classes. However, the confidence level and recognition ability of our model for the detection of unknown objects still need to be improved, and this is what we will strive for in the future. Currently, our model still faces the issue of mistakenly detecting the background as unknown objects, and the benchmark we proposed is not yet comprehensive, including only a single-moment task.\par
In our future work, these two points will be the main focus of our research. In addition, post-processing operations for prediction boxes of unknown objects are urgently needed, so developing a post-processing operation similar to NMS, dedicated to unknown objects, is also a direction that requires our research. To facilitate the integration of open-world object detection algorithms into everyday use and contribute positively to society, we are fully committed to diligently pursuing this goal with great care and effort.

\section{Conclusions}
In this paper, we start by observing that the simple \textbf{knowledge distillation} could convert the open-world knowledge in the large pre-trained vision-language grounding model for the specialized OWOD task and propose a simple framework with surprisingly good performance. Meanwhile, we propose the \textbf{down-weight training loss function} for the detector's mixed learning process of known ground truth, distilled unknown knowledge, and the pseudo unknown knowledge in OWOD algorithm to mitigate the effect of distilled knowledge on the detection performance of known objects. Besides, the \textbf{cascade decoupled detection transformer structure} is proposed to alleviate the influence caused by unknown objects on detecting known objects. Last but not least, we propose two novel benchmarks to comprehensively evaluate the ability of the open-world detectors to detect unknown open-world objects, named \textbf{StandardSet}$\heartsuit$ and \textbf{IntensiveSet}$\spadesuit$ respectively, based on the complexity of their testing scenarios. Our extensive experiments on existing and proposed benchmarks demonstrate the effectiveness of our framework. Our model exceeds the distilled large pre-trained vision-language grounding model for OWOD and state-of-the-art methods for OWOD and IOD. 

\ifCLASSOPTIONcompsoc
  \section*{Acknowledgments}
\else
  \section*{Acknowledgment}
\fi

This work is supported by National Natural Science Foundation of China (grant No.61871106 and No.61370152), Key R\&D projects of Liaoning Province, China (grant No.2020JH2/10100029), and the Open Project Program Foundation of the Key Laboratory of Opto-Electronics Information Processing, Chinese Academy of Sciences (OEIP-O-202002).

\ifCLASSOPTIONcaptionsoff
  \newpage
\fi

\bibliographystyle{ieeetr} 
\bibliography{reference}

\begin{IEEEbiography}
[{\includegraphics[width=1in,height=1.25in,clip,keepaspectratio]{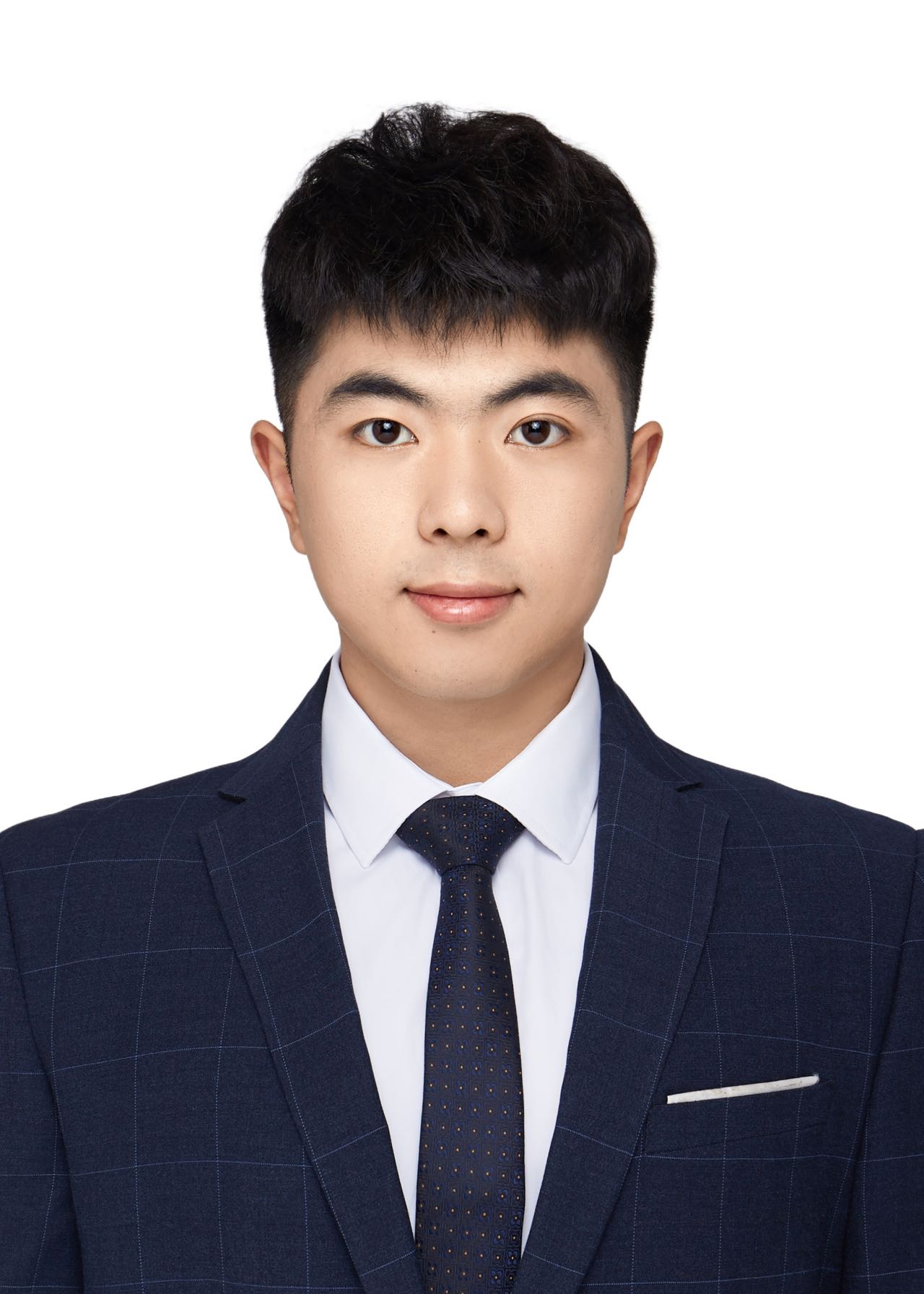}}]{Shuailei~Ma}
received the B.S. degree from Northeastern University, Shenyang, China, in 2022. Currently, he is pursuing a Ph.D. degree in the College of Information Science and Engineering at Northeastern University. His research interests include computer vision and deep learning.
\end{IEEEbiography}

\begin{IEEEbiography}
[{\includegraphics[width=1in,height=1.25in,clip,keepaspectratio]{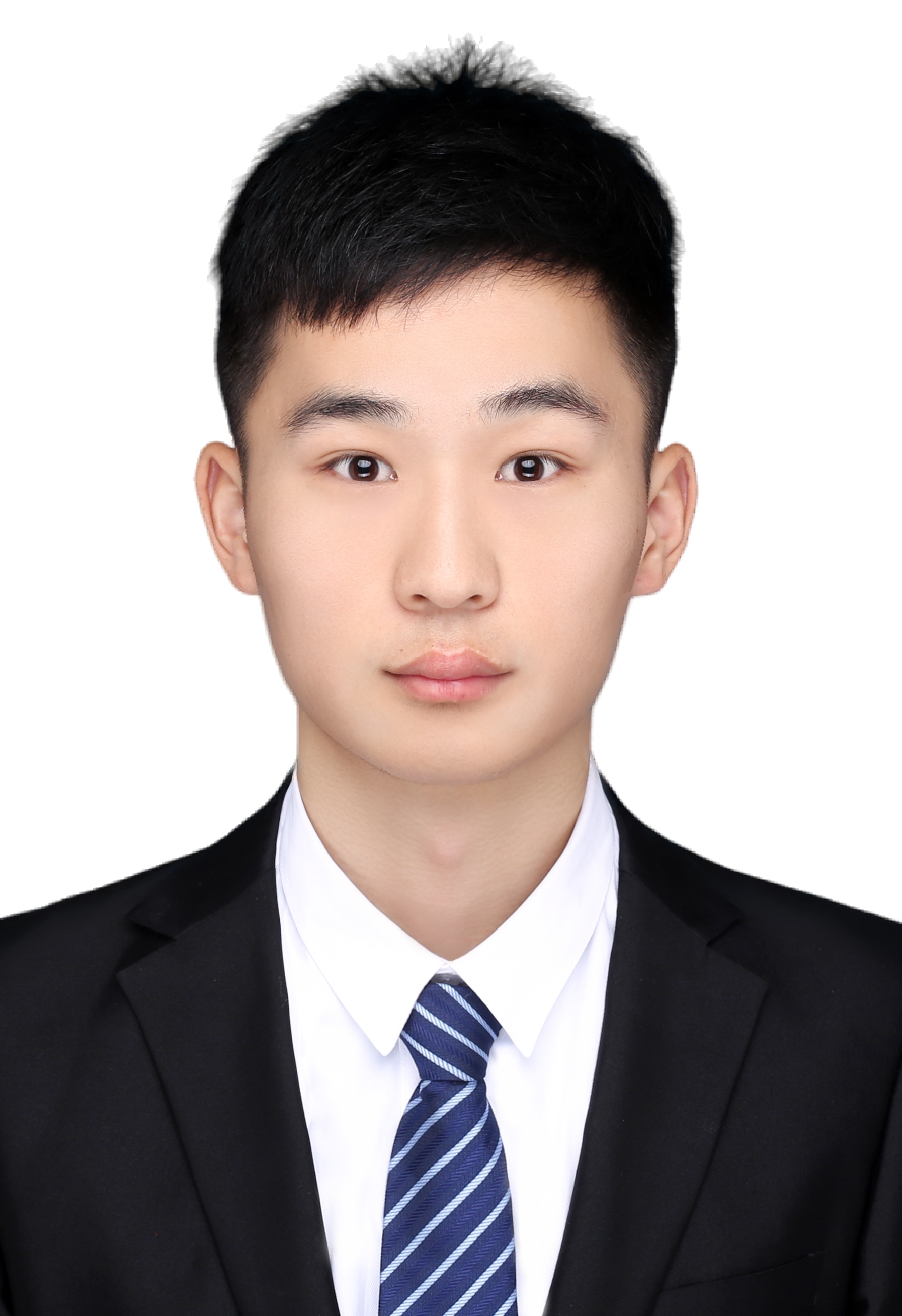}}]{Yuefeng~Wang}
received the B.S. degree from Nanjing University of Information Science and Technology, Nanjing, China, in 2021. Currently, he is pursuing an M.Sc. degree in the College of Information Science and Engineering at Northeastern University. His research interests include computer vision and deep learning.
\end{IEEEbiography}

\begin{IEEEbiography}
[{\includegraphics[width=1in,height=1.25in,clip,keepaspectratio]{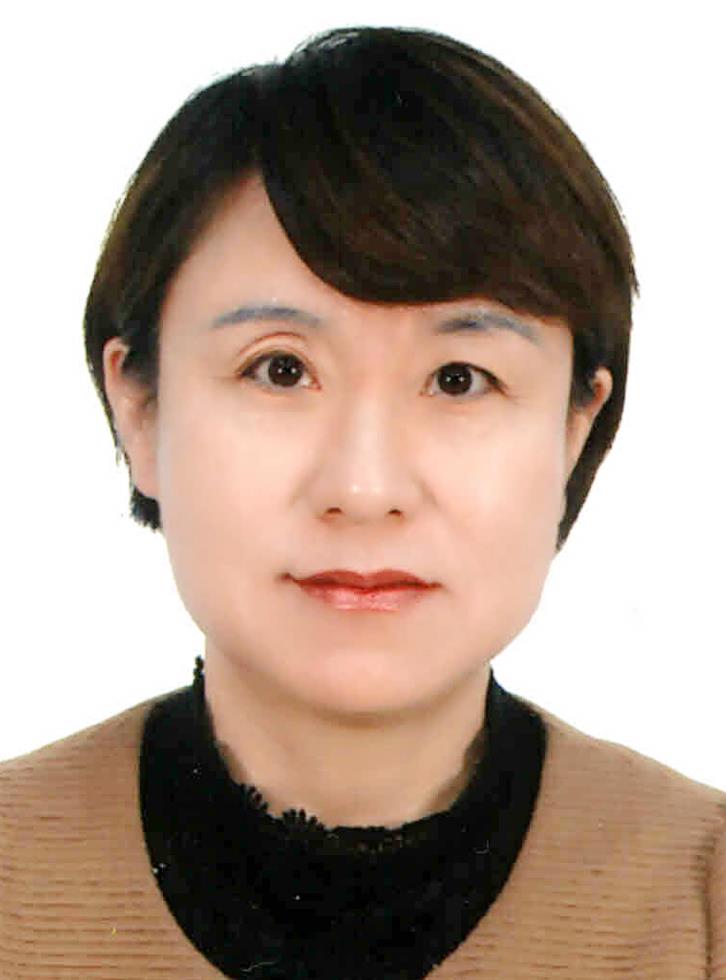}}]{Ying~Wei}
received the B.Sc. degree from Harbin Institute of Technology, China in 1990, and received the M.Sc. and the Ph.D. degree from Northeastern University, China in 1997 and 2001, respectively. Her research interests include image processing \& pattern recognition, computer vision, medical image computation and analysis, and deep learning, etc. She is now a full-time professor at Northeastern University, China. She has more than 60 journal papers and 5 granted patents in her research fields.
\end{IEEEbiography}

\begin{IEEEbiography}
[{\includegraphics[width=1in,height=1.25in,clip,keepaspectratio]{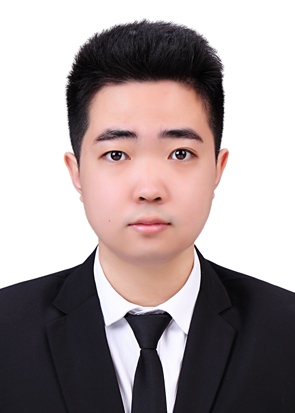}}]{Jiaqi~Fan}
received a B.S. degree from Northeastern University, China, in 2022. Currently, he is pursuing an M.Sc. degree in the College of Information Science and Engineering at Northeastern University. His research interests include computer vision and deep learning.
\end{IEEEbiography}

\begin{IEEEbiography}
[{\includegraphics[width=1in,height=1.25in,clip,keepaspectratio]{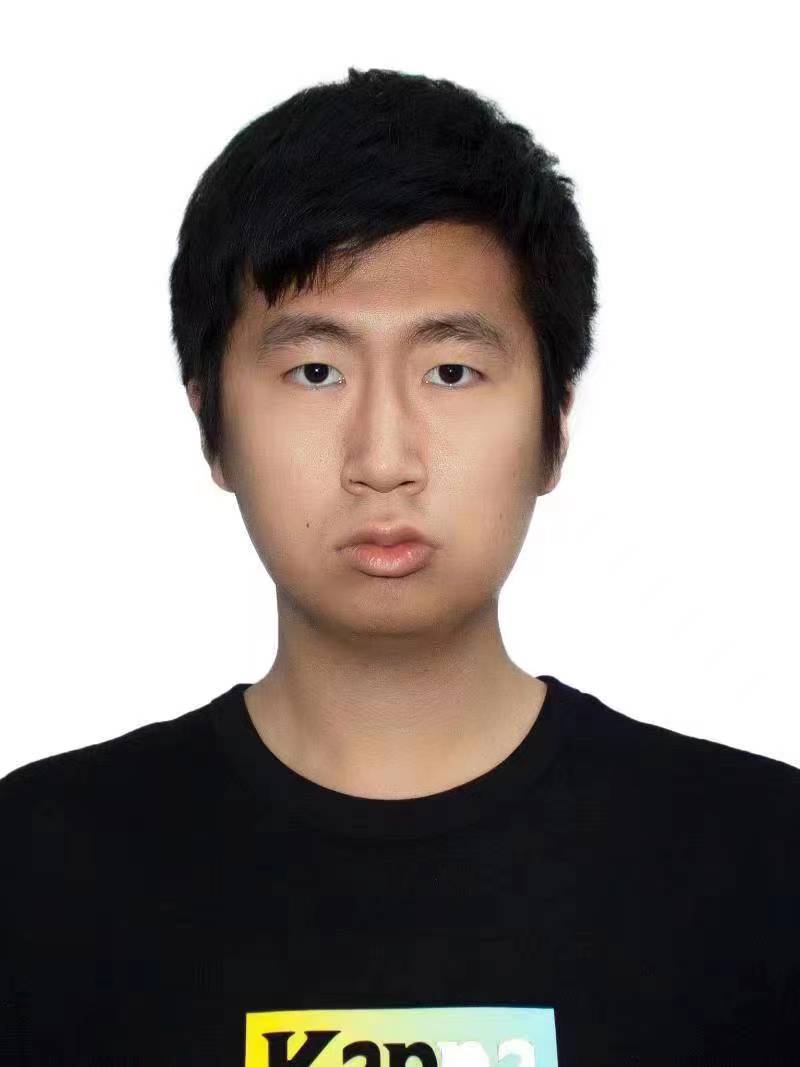}}]{Enming~Zhang}
received a B.S. degree from Northeastern University, China, in 2022. Currently, he is pursuing an M.Sc. degree in the College of Information Science and Engineering at Northeastern University. His research interests include computer vision and deep learning.
\end{IEEEbiography}

\begin{IEEEbiography}
[{\includegraphics[width=1in,height=1.25in,clip,keepaspectratio]{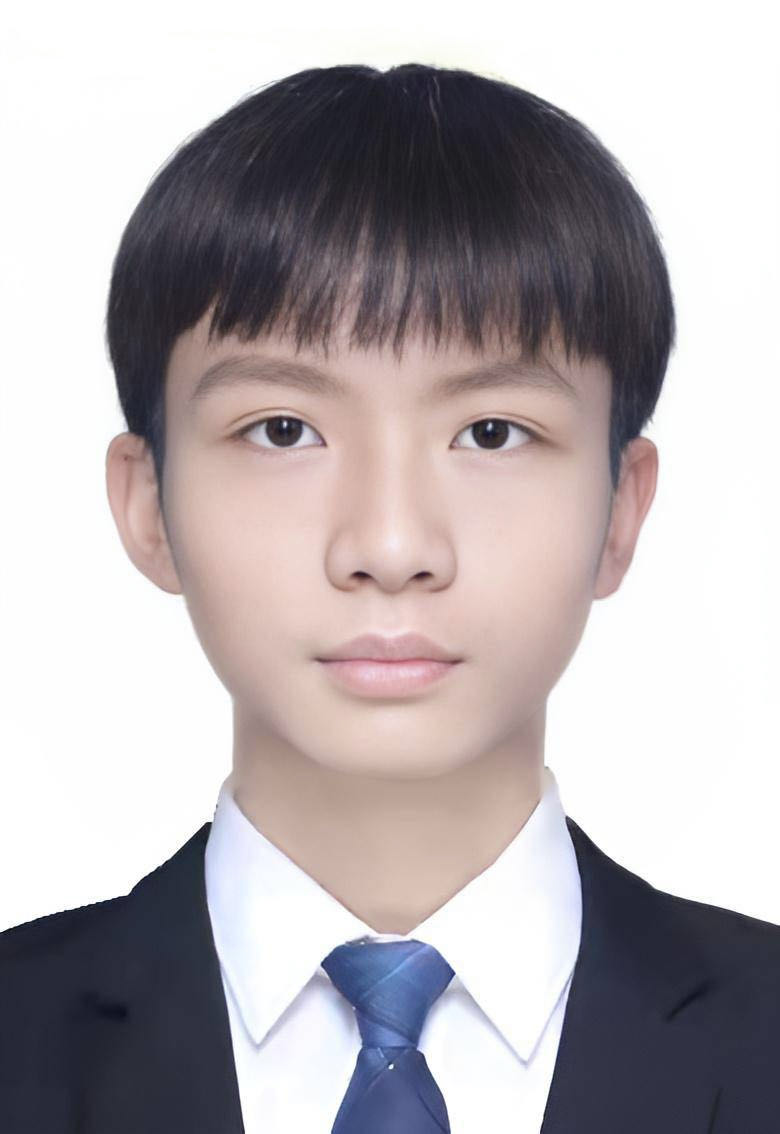}}]{Xinyu~Sun}
received the B.E. degree in Automation Science and Engineering from South China University of Technology, China, in 2021. He is working toward the M.Sc. degree in the School of Software Engineering, South China University of Technology, China. His research interests include embodied AI and multi-modal video understanding.
\end{IEEEbiography}

\begin{IEEEbiography}
[{\includegraphics[width=1in,height=1.25in,clip,keepaspectratio]{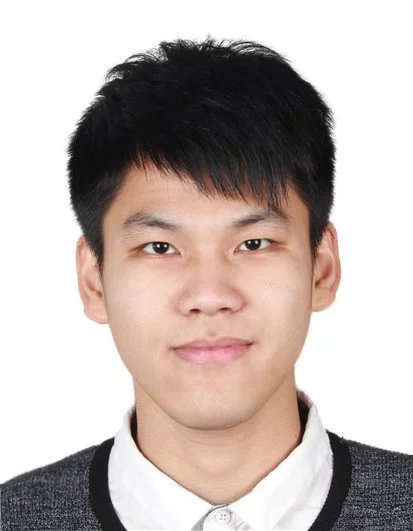}}]{Peihao~Chen}
received the B.E. degree in Automation Science and Engineering from South China University of Technology, China, in 2018. He is working toward the PhD degree in the School of Software Engineering, South China University of Technology, China. His research interests include embodied AI and multi-modal video understanding.
\end{IEEEbiography}

\end{document}

%% file: def.tex
\def\etal{{\em et al.}}

\DeclareMathAlphabet\mathbfcal{OMS}{cmsy}{b}{n}

\def\0{{\bf 0}}
\def\1{{\bf 1}}

\def\eg{\emph{e.g.}} 
\def\ie{\emph{i.e.}}

\def\etal{\emph{et al.}}

\usepackage{amsmath}